\pdfoutput=1

\documentclass[11pt]{article}

\usepackage[final]{mtsummit25}
\usepackage{copyright}

\usepackage{times}
\usepackage{latexsym}

\usepackage[T1]{fontenc}

\usepackage[utf8]{inputenc}

\usepackage{microtype}

\usepackage{inconsolata}

\usepackage{graphicx}
\usepackage{multicol}
\usepackage{multirow}
\usepackage{pdflscape}
\usepackage{tabularx}
\usepackage{makecell} 
\usepackage{amsmath}

\title{To MT or not to MT: An eye-tracking study on the reception by Dutch readers of different translation and creativity levels }

\author{Kyo Gerrits and Ana Guerberof Arenas \\
  Center for Language and Cognition (CLCG), University of Groningen \\
  \texttt{k.gerrits@rug.nl} \and {a.guerberof.arenas@rug.nl}} 

\begin{document}
\maketitle
\begin{abstract}
This article presents the results of a pilot study involving the reception of a fictional short story translated from English into Dutch under four conditions: machine translation (MT), post-editing (PE), human translation (HT) and original source text (ST). The aim is to understand how creativity and errors in different translation modalities affect readers, specifically regarding cognitive load. Eight participants filled in a questionnaire, read a story using an eye-tracker, and conducted a retrospective think-aloud (RTA) interview. The results show that units of creative potential (UCP) increase cognitive load and that this effect is highest for HT and lowest for MT; no effect of error was observed. Triangulating the data with RTAs leads us to hypothesize that the higher cognitive load in UCPs is linked to increases in reader enjoyment and immersion. The effect of translation creativity on cognitive load in different translation modalities at word-level is novel and opens up new avenues for further research. All the code and data are available at \url{https://github.com/INCREC/Pilot_to_MT_or_not_to_MT}.
\end{abstract}

\section{Introduction}
Recently, publishing houses have been more vocal about the use of machine translation (MT) in their translation process \citep{klemin}, arguing that the output quality is good enough to post-edit certain genres considered less literary such as crime or romance novels in certain language combinations. However, no data has been provided to illustrate not only the impact on translators' livelihood and sustainability of high-quality literary translations but also what the impact on the readers of these books might be. Our research focuses on the latter, i.e. we seek to explore the effects of the use of MT-mediated texts in literary translation.

Having this goal in mind, this study uses materials and research methods from an existing study by \citet{Guerberof2024} that explored how different reading modalities (MT, PE, HT and ST) affect Dutch readers regarding engagement, enjoyment, and translation reception. We include two new methodological parts: a) an eye-tracking device to obtain granular data on readers' attention (cognitive load) and b) retrospective think-aloud (RTA) interviews to understand the differences in the readers' experiences when reading these texts. We  focus on the effect of creativity across the different modalities by analysing the reception of units of creative potential. A unit of creative potential is a word or group of words that present a problem to the translator that requires a higher level of creativity, see Section 2.2

With this experiment, our main aim is to find a methodological framework to measure cognition and creativity, which, to our knowledge, has not been attempted before. To test this methodology, we guide the experiment with the following research questions: 
RQ1: Do readers have a higher cognitive load in units of creative potential than in other regular parts of the sentence? 
RQ2: Do readers process these units differently according to the translation modality? 
RQ3: Do readers process the translators' solutions differently according to the level of creativity?  
RQ4: Do errors in a segment increase the cognitive load of the reader?

\section{Reading and translation reception}
\subsection{Eye-tracking and reading studies}
Eye-tracking is a common method for measuring cognitive load. Even before technology was used for measuring the eye-movements, researchers already thought cognitive effort influenced eye movements during reading—-now known as the cognitive-control hypothesis \citep{RAYNER2015107}. Yet there were also doubts about eye-tracking methods. As average fixations are around 250 ms, some thought that it would be too little time for lexical processing \citep{Chanceux}. Others also believed eye movements were largely caused by involuntary oculomotor movements \citep{Yarbus}. However, numerous studies have since shown that eye movement is in fact heavily influenced by cognitive effort (\citealp{Reichle2013}; \citealp{Schotter2017}; \citealp{AlMadi2020}; \citealp{Dias2021}).

Many reading studies focus on comprehension; although intuitively we might think that lower reading times correlate with better comprehension, research shows that increased fixation duration and count might indicate higher comprehension levels (\citealp{Meziere2023}; \citealp{Southwell2020}; \citealp{Wonnacott2016}). 
Furthermore, studies looking at reading effort in literary texts find the more literary and style-heavy a text is considered to be, the more and longer the fixations are (\citealp{Corcoran2023}; \citealp{Fechino2020}), especially foregrounded elements (\citealp{Jacobs2015}; \citealp{Muller2017}).\footnote{Stylistic devices that emphasize certain parts of the text to increase the impact on the reader.} \citet{Torres2021} also found that this increase in cognitive load also seems related to increased immersion in and engagement with the text, as reported by participants. 

\subsection{Creativity in translation}
\begin{figure}[h]
    \centering
    \includegraphics[scale=0.33]{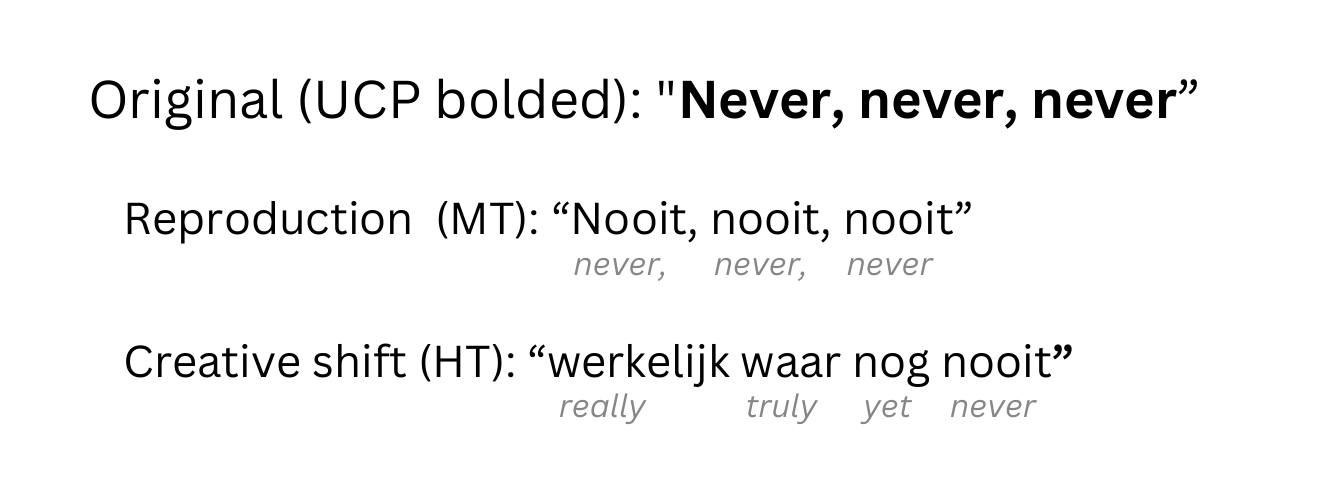}
    \caption{An example of UCP, Reproduction, and CS from the experiment, including word-level glosses.}
    \label{fig:UCP1}
\end{figure}

A clear conceptualisation of creativity in translation studies is introduced by \citeauthor{Kussmaul1991} in his seminal work (\citeyear{Kussmaul1991}; \citeyear{Kussmaul1995}; \citeyear{Kussmaul2000a}; \citeyear{Kussmaul2000b}). This is further operationalised by \citeauthor{Bayer-Hohenwarter2009} (\citeyear{Bayer-Hohenwarter2009}; \citeyear{Bayer-Hohenwarter2010}; \citeyear{Bayer-Hohenwarter2011}; \citeyear{Bayer-Hohenwarter2013}). They describe a creative translation as “involv[ing] changes (…) when compared to the source text, thereby bringing something that is new and also appropriate” (\citeauthor{Bayer-Hohenwarter2020}, \citeyear{Bayer-Hohenwarter2020}, p. 312). \citet{Guerberof2020} further develops the concepts of unit of creative potential (UCP) and creative shift (CS) to measure creativity in literary translation and see the impact creativity has on readers. UCPs are problematic units in the source text that translators cannot translate routinely and for which they have to use problem-solving abilities, that is, their creative skills \citep{Bayer-Hohenwarter2011}. CSs are translated UCPs in which the translation deviates from the original, in contrast to Reproductions, where UCPs do not deviate from the original in structure or where there is already a coined translation \citep{Guerberof2020}. This is illustrated in Figure \ref{fig:UCP1}. Our study uses this conceptualisation of creativity to annotate the translations, which allows us to analyse cognitive load (eye-tracking) across creativity (RQs 1, 2 \& 3).

\subsection{Reading in the Netherlands}
Dutch readership has some peculiarities worth mentioning regarding the cohabitation of English and Dutch languages. Recent market research shows that sales of foreign language books have increased 124\% since 2020, accounting for 25\% of all sales in 2024 \citep{KVBBoekwerk_2025}, with the majority of these being English. Many readers read or even prefer reading books in the original language: 41\% of Dutch readers regularly read books in another language, of which 77\% are books in English; 29\% even prefer English-language books to Dutch ones \citep{KVBBoekwerk_2022}. This is interesting both for the potential influence of English phrases or structures on readers, and because English might be their default when opting to buy a book. 

\subsection{Literary translation and MT}
A known issue in literary MT is the presence of errors, despite continuous improvements (\citealp{stasimioti-etal-2020-machine}, \citealp{matusov-2019-challenges}). Recent developments in neural machine translation (NMT) and large language models (LLMs) show to have increased MT quality \citep{SonKim2023}, but studies looking at LLMs for literary translation still find numerous errors in the MT output \citep{zhang2024goodllmsliterarytranslation}, even when prompted to correct this MT output (\citealp{egdom-etal-2024-make}; \citealp{macken-2024-machine}). In the particular case of English-Dutch MT: \citet{fonteyne-etal-2020-literary} and \citet{tezcan-etal-2019-sport} look at an English-to-Dutch NMT version of Agatha Christie's \textit{The Mysterious Affair at Styles} and find that 44\% sentences had no errors while 56\% still contain errors. A follow-up study by \citet{Webster2020} looks at four different novels and finds a much higher number of incorrect sentences in the NMT output, 77\% with errors vs 23\% without. They argue that this difference could be related to the ST linguistic complexity. 

If PE is considered, some studies indeed show a decrease in errors when compared to raw MT output (\citeauthor{Guerberof2020}, \citeyear{Guerberof2020}; \citeyear{Guerberof2022}). PE might also provide a lower cost for industry or shorten the translation times \citep{Toral2018}, although this also depends on the desired final quality. However, PE is not without challenges. One of these is that the MT output tends to prime post-editors and this results in a final text that is syntactically, semantically and stylistically closer to MT than to HT in both technical and literary texts (\citealp{toral-2019-post}; \citealp{daems-etal-2024-impact}; \citealp{macken-etal-2022-literary}; \citealp{Kolb2021}; \citealp{castilho-resende-2022-mt}). PE has also been shown to reduce literary style and authorial voice compared to HT (\citealp{KennyWinters}; \citealp{Mohar2020}; \citealp{sahin-gurses-2019-mt}). Last but not least, translators have expressed their dislike of using PE, preferring to translate from scratch for creative purposes (\citealp{Moorkens2018}; \citealp{Daems2019}). 

\subsection{Translation reception and MT}
Translation reception concerns how readers react to a translation, such as emotional (e.g. enjoyment) and cognitive responses (confusion, reading times). There are relatively few studies on translation reception \citep{Walker2021}. Some studies focus on the effect of errors in non-literary MT: \citet{Kasperaviciene2020} and \citet{stymne-etal-2012-eye}, for instance, use eye-tracking to analyse the effect of errors in newspaper articles and political discussions respectively. They find that total fixation duration and fixation count are higher on sections that contain errors. \citet{Whyatt2024}, also using an eye-tracker, analyse the reception of newspaper translations of low and high quality and find that participants spend more time on sentences of lower quality and sentences with errors than in those without. These studies suggest that a text with more errors and of lower quality garners more cognitive load, at least in non-literary texts.

Others look at literary MT. \citet{colman-etal-2022-geco}, for instance, explore the reception of an English-into-Dutch MT version of Agatha Christie's \textit{The Mysterious Affair at Styles} and compare it to a published translation. 20 participants were eye-tracked while reading the entire novel, alternating MT or HT every 25\%. They find lower readability of MT compared to the published translation.

\citeauthor{Guerberof2020} (\citeyear{Guerberof2020}; \citeyear{Guerberof2024}) consider the literary reception of MT, PE, HT and ST from a creativity angle. Their reception studies look at reader responses to the different modalities using narrative engagement, enjoyment, and translation reception scales. They find that reading experience is not significantly different in HT and PE, but MT scored significantly lower on the three variables. However, results differ per text and language; in Dutch, for example, ST scores higher than PE and HT, which was not the case for Catalan readers. Our study builds upon this study as mentioned above, see Section 3.1 for details.

\section{Measuring creativity and cognitive load}
This section describes the data, annotation criteria, participants, eye-tracking device, questionnaire \& RTA interviews, as well as the preprocessing of the initial dataset and statistical modelling. Combining quantitative and qualitative methods allows us to triangulate the data to understand the complex interactions of variables, such as creativity and cognitive load, to answer our RQs.

\subsection{Content}
The methodology of this study borrows the open dataset containing the original and translated texts, the annotations, and the questionnaire (on engagement, enjoyment and reception) from \citet{Guerberof2024}.

The text, the science-fiction short story "2BR02B" (1962) by Kurt Vonnegut, contains 123 paragraphs, 234 sentences and 2548 words. Two professional translators created the HT and PE versions: to ensure readers would not rate a text higher due to translator preference, each translator did half of the text without MT (HT) and the other half with MT (PE). The MT was created by a customized NMT engine trained on literary texts, based on transformer architecture \citep{Toral2023}.

The ST was annotated by two professional translators. They identified 185 UCPs in the English ST. Subsequently, two professional reviewers annotated the Dutch TTs for creativity and errors. The UCPs in the Dutch TTs were annotated as either CS, Reproduction (no CS), NAs (too many errors for classification) or omissions (UCP is omitted entirely). Errors were classified and annotated using the harmonised DQF-MQM Framework,\footnote{\url{https://themqm.org/error-types-2/typology/}} which categories the type of error and its severity. For more details on these annotations, see \citet{Guerberof2022}.The total number of CSs, Reproductions, and errors is shown in Table \ref{table:1}, including a creativity index (CI), which combines CSs and errors according to the following formula (in \citet{Guerberof2022}):

\vspace{1mm}

$\text{CI} = (\frac{\#CS}{\#UCPs} - \frac{\#ErrorPoints-\#Kudos}{\#Words\_in\_ST}) * 100 $

\vspace{2mm}

Table \ref{table:1} shows that for the texts we are using in this experiment, the CI for HT is the highest followed by PE and lastly by MT.

\begin{table}[h!]
\centering
\begin{tabular}{ l  l  l  l }
\hline
 \textbf{Creativity} & \textbf{HT} & \textbf{PE} & \textbf{MT} \\ 
\hline
 Creative shifts & 79 & 63 & 26\\
 Reproduction & 105 & 122 & 143 \\
 Errors & 75 & 221 & 528 \\
 \textbf{Creativity Index} & \textbf{41} & \textbf{25} & \textbf{-7} \\
\hline
\end{tabular}
\caption{Results of the error and UCP annotation.}
\label{table:1}
\end{table}

\subsection{Participants}
Eight participants (six women and two men) were recruited who voluntarily signed up using a Google form from a flyer distributed throughout universities and were paid €20 after completing the experiment. The criteria for selecting participants were to be native Dutch speakers (1) and frequent readers (2), reading at least one book per month. Five had a master's degree and three had a BA or were in the process of graduating. Participants were randomly assigned to one of the four modalities to read, resulting in two participants per modality. The experiment was reviewed by the Ethics Committee at the University of Groningen, and participants gave their written informed consent.

\subsection{Eye-tracking}
To gather data on cognitive load, we employ an EyeLink Portable Duo eye tracker (SR Research), combined with a 27-inch monitor with a resolution of 1024 x 768 pixels. Participants used a headstand for optimal sampling rate (2000Hz), with the headset set at a 55 cm distance to the eye tracker. The experiment was set up using EyeLink’s Experiment Builder and carried out at the EyeLab of Groningen between May 1st and May 15, 2024. Participants were calibrated and validated using a nine-point grid, and participants were recalibrated if the calibration or validation was poor or when the deviations in validation were above 0.5. Interest areas were automatically created by Experiment Builder based on word boundaries, so that each word was a separate AOI. The text was presented on the monitor, and the font (Arial) and font-size (35) were chosen for readability (\citealp{Minakata2021}; \citealp{Masulli2018}). Each screen contained about 10 lines of text and participants could move to the next page themselves (by clicking or pressing a random key), which would be preceded every time by a drift check in the upper-left corner—the same place the first word of the new page would appear. A pause was included in the middle of the story, after about 1310 words which tended to be about 15 minutes of reading, where the original text also had a separating dinkus. After the pause, calibration and validation were repeated. There were no time constraints for the participants. 

To answer our four research questions, we focused on five dependent variables: total fixation duration (TFD, duration of all fixations), first-pass time (FPT, duration before word is first exited), regression path (RP, duration before word is first exited to the right), fixation count (FC, number of fixations) and regression count (RC, number of regressions, which are fixations from words that come after the word in question). Previous eye-tracking studies have shown the relevance of these variables for measuring cognitive load \citep{Skaramagkas}, and for translation specifically \citep{Vanroy2022}. However, it is not always clear which specific measures will be the most appropriate for our RQs. Regression is, for instance, often associated with confusion, but also with increased comprehension and skim reading \citep{Southwell2020}. Nevertheless, TFD is often seen as a general indication of cognitive load, FPT for immediate and semantic understanding, with RP, FC and RC for contextual understanding. As we are interested in cognitive load, our main variable of interest is TFD. Due to space constrains, the results for the other dependent variables are shown in Appendix \ref{sec:appendix-ET}. 

\subsection{Questionnaire}
Although our main focus was on cognitive load, we also wanted to gather the demographics and reading habits of the participants, as well as their engagement, enjoyment, and reception of translation. Therefore, we also used the questionnaire from the previous on-line experiment \citep{Guerberof2024}. In a computer in the lab, participants completed firstly the sections on demographics and reading patterns, and secondly, after the eye-tracking experiment, participants completed the comprehension (10 items), narrative engagement (15 items), enjoyment (3 items) and translation reception (9 items) parts. The comprehension questions were multiple-choice, while the other three used a 7-point Likert scale.\footnote{Questionnaire can be found in Appendix \ref{sec:appendix-Q} and on GitHub.}

\subsection{Retrospective think-aloud interview}	
\begin{figure}[h]
    \centering
     \includegraphics[scale=0.36]{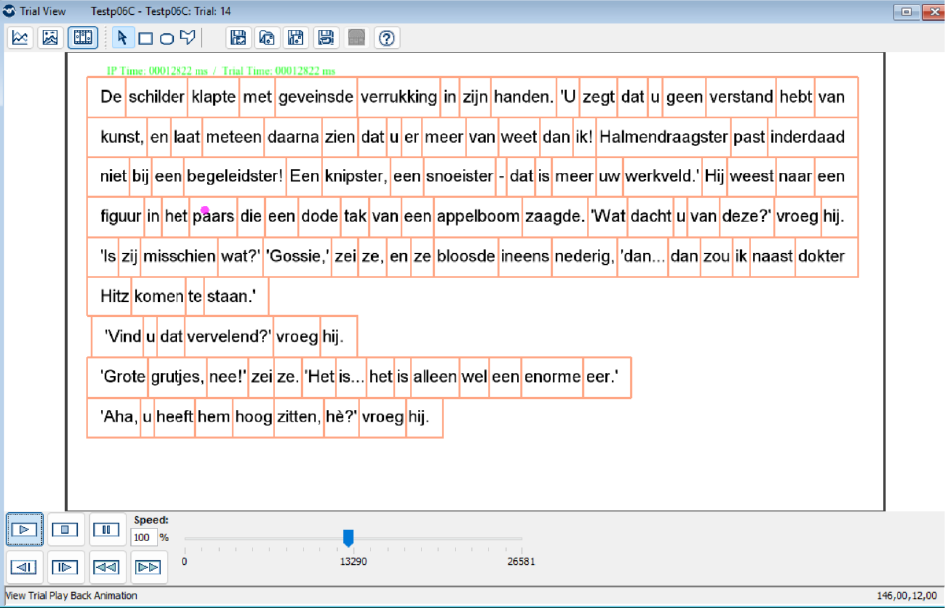}
    \caption{the Trial Play Back Animation feature in which the (moving) purple dot replays the gaze movements in real time.}
    \label{fig:Data_viewer}
\end{figure}

Immediately after finishing the eye-tracking experiment and completing the questionnaires, an RTA protocol took place to triangulate the data and better understand our eye-tracking results.
Participants were prompted with visualisations of their gaze during reading, but were free to discuss or mention any aspect of the story. Visualisations were created automatically with Data Viewer, using its Trial Play Back Animation feature, which shows a participant’s gaze as it moves over the text in real-time, see Figure \ref{fig:Data_viewer}. This allows participants not only to comment on the text but also by seeing their eye movements clarify things they seemed to pause at or go back to. The interviews were conducted by one of the researchers in English for processing ease, except in three instances where the participants preferred to speak in Dutch; these are translated for analysis by the same researcher. 

\subsection{Dataset and preprocessing}
\begin{figure}[h]
    \centering
     \includegraphics[scale=0.4]{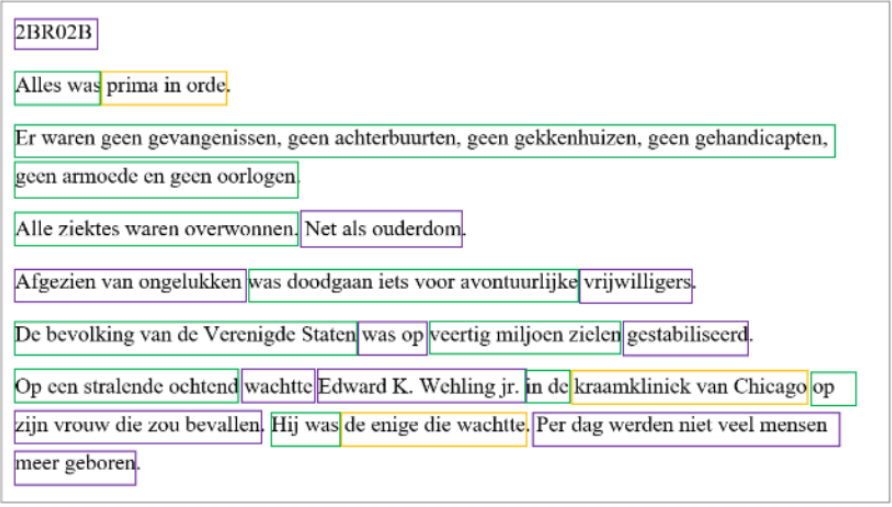}
    \caption{Trial of PE version, divided in segments: entire sentences or phrases. UCPs are yellow (CS) or purple (Rep.), and non-UCP segments are green.}
    \label{fig:units}
\end{figure}

The data was processed using Data Viewer; 20624 observations were generated corresponding to each word in the texts, the default AOIs. These observations were further classified using the existing annotations, e.g. if a word was part of CS or Reproduction or if it was part of an error. This is our dataset I, i.e. containing data per word. Since we wanted to compare the eye-tracking data in the different translated texts for each UCP in the ST, we created a second dataset, (n=3618). In dataset II each observation is a segment either containing a UCP or without a UCP. In this way, we could compare the translation solutions for the 185 UCPs in the three translation modalities, illustrated in Figure \ref{fig:units}. As these segments have different numbers of words, we normalised the dependent variables from the eye-tracker according to words per segment. We primarily used this dataset II for our analyses as this dataset was better suited to answer our RQs which deal with UCPs, CSs and Reproductions.
We use dataset I (per word) to check word frequency and descriptive results, see Appendix \ref{sec:appendix-ET}.

\

\subsection{Statistical modelling for eye-tracking data}
For the statistical analysis of our eye-tracking data, we used a regression model to analyse the main effects and the interactions of our independent variables Modality (HT, PE, MT and ST), Creativity (CSs, Reproductions, omissions and NAs) and Errors (with or without errors). After analysing the data, we fitted a generalized additive mixed effect model (GAM model), using scaled t-distribution. We decided on this model as our data was not normally distributed, even after log-transformations. This was partly due to zero measurements (sections without any fixations), so we only use the data containing fixations to create the model and analyse the zeros independently.\footnote{The analysis of the zero measurement can be found in Appendix \ref{sec:appendix-ET}.} We used the eye-tracking measures (TFD, FPT, RP, FC, RC) as dependent variables, with Modality, Creativity and Error as independent variables. These were contrast coded, as well as contrasting HT against PE specifically to study the difference between those two in more detail, as previous studies by \citeauthor{Guerberof2020} (\citeyear{Guerberof2020}; \citeyear{Guerberof2024}) showed little difference in reception between them. The random effects were the participants and the UCPs. 

The model's explained variance was about 40\% for the dependent variable TFD ($R^{2}$ = 0.379), but only 28\% ($R^{2}$ = 0.241) and 27\%  ($R^{2}$ = 0.241) for FPT and RP, while FC and RC did not meet the required assumptions for this model. Therefore, we performed non-parametric tests for those four of the eye-tracking measures (FPT, RP, FC \& RC).

\subsection{Analysing frequency}
We also wanted to check the effect of word frequency on our data, since research has shown a strong inverse link between the frequency of words and cognitive load (\citealp{Schilling1998}; \citealp{Holmqvist2011}; \citealp{Walker2021}), and UCPs contain words or expressions that are less frequent.  We extracted the word frequencies for all Dutch words, using the \textit{wordfreq} Python library \citep{robyn_speer_2022_7199437}. This library is based on the Exquisite Corpus, which is a multilingual corpus compiled of eight different domains of text. The ‘best' wordlist, available for Dutch, includes words that appear at least once per 100 million words, making it a reliable corpus. For this particular analysis, dataset I was used because word frequencies are calculated on a word-level.

\section{Cognitive load, creativity and reading experience}
In this section, we discuss the results from the questionnaires, eye-tracking device and RTAs that serve to answer our four RQs.
\subsection{Self-reported user experience}
Table \ref{tab:Part} shows the results from the questionnaire regarding participants' demographic information and reading habits, which shows similar patterns across the participants.  

\begin{table}[h!]
    \centering
    \small
    \begin{tabular}{l l l l l}
    \hline
    \textbf{Modality} & \textbf{n} & \textbf{Age }  & \textbf{Reading habit} \\
    \hline
    PE & 2 & 18 - 24 & 4.5 \\
    MT & 2 & 18 - 34 & 4 \\
    HT & 2 & 18 - 24 & 4.5\\
    ST & 2 & 25 - 34 & 4.5 \\
    \hline
    \end{tabular}
    \caption{Participants' age and reading habits from the participants (1 = Never, 2 = Once per 3 months, 3 = Once per month, 4 =  Multiple times a week, 5 = Daily).}
    \label{tab:Part}
\end{table}

Table \ref{table:questionnaire} shows the results from the questionnaire for comprehension, narrative engagement, enjoyment and translation reception from the 8 participants. Participants rated MT the lowest across all scales. HT and PE are rated higher than MT, with HT scoring higher than PE on narrative engagement and enjoyment, but not on translation reception. There are two interesting results: ST scores lower than both HT and PE on narrative engagement and reception, although not in enjoyment; and MT has the highest mean score for comprehension. Despite MT scoring the highest in the multiple-choice comprehension, the RTAs show that participants did not enjoy reading MT and reported struggling to understand the narrative (see Section 4.3). This could mean that readers in MT understand the basic details of the story so they can respond to basic questions and that they compensate using the context when they do not understand certain elements in MT; this strategy of compensation to understand MT has been reported in previous studies \citep{GuerberofMoorkens2023}.

\begin{table}[h!]
\small
\centering
\begin{tabular}{ l l l l l l l}
\hline
\multirow{1}{*}{\textbf{Mod.}} & \textbf{Category} & \textbf{n} & \textbf{Mean} & \textbf{SD} & \textbf{Med.} & \textbf{Min} \\
\hline
\multirow{4}{*}{HT} & Comp. & 2 & 8.5 & 0.71 & 8.5 & 8  \\  
& Eng. & 30 & 5.37 & 1.35 & 6 & 2  \\
& Enj. & 6 & 4.67 & 1.21 & 4.5 & 3  \\
& T.R. & 18 & 4.39 & 1.42 & 5 & 2  \\
\hline
\multirow{4}{*}{MT} & Comp. & 2 & 9.5 & 0.71 & 9.5 & 9  \\
&  Eng. & 30 & 3.63 & 1.47 & 4 & 1  \\
& Enj. & 6 & 2.17 & 1.47 & 2 & 1  \\
& T.R. & 18 & 2.06 & 1.11 & 2 & 1  \\
\hline
\multirow{4}{*}{PE} & Comp. & 2 & 6.5 & 0.71 & 6.5 & 6  \\
& Eng. & 30 & 5.13 & 1.48 &5.5  &  2 \\
& Enj. & 6 & 4.17 & 0.75 & 4 & 3 \\
& T.R. & 18 & 5.06 & 1.06 & 5 & 3  \\
\hline
\multirow {4}{*}{ST} & Comp. & 2 & 9 & 1.41 & 9 &  8  \\
& Eng. & 30 & 4.20 & 1.58 & 5 & 1  \\
& Enj. & 6 & 4.67 & 0.82 & 4.5  & 4 \\
& T.R. & 18 & 4.22 & 1.63 & 4 &2   \\
\hline
\end{tabular}
\caption{Results of the questionnaire per modality and scale (comprehension (10 multiple-choice questions), narrative engagement, enjoyment and translation reception (each 7-point Likert scale).}
\label{table:questionnaire}
\end{table}

These results correspond moderately with the results from \citet{Guerberof2024}: there, MT scored lowest, and ST ranked highest for narrative engagement and enjoyment. We have identified two potential causes for the difference in this second experiment: the most obvious one is that here we only had two participants per modality, which does not allow for generalization; the other reason could be that in our study, participants read the text in a lab using an eye-tracker, which could indicate higher levels of attention as opposed to reading online at home as in the original experiment.

\subsection{Cognitive load}
\subsubsection{Descriptive results}
 
\begin{figure}[h]
    \centering
     \includegraphics[scale=0.38]{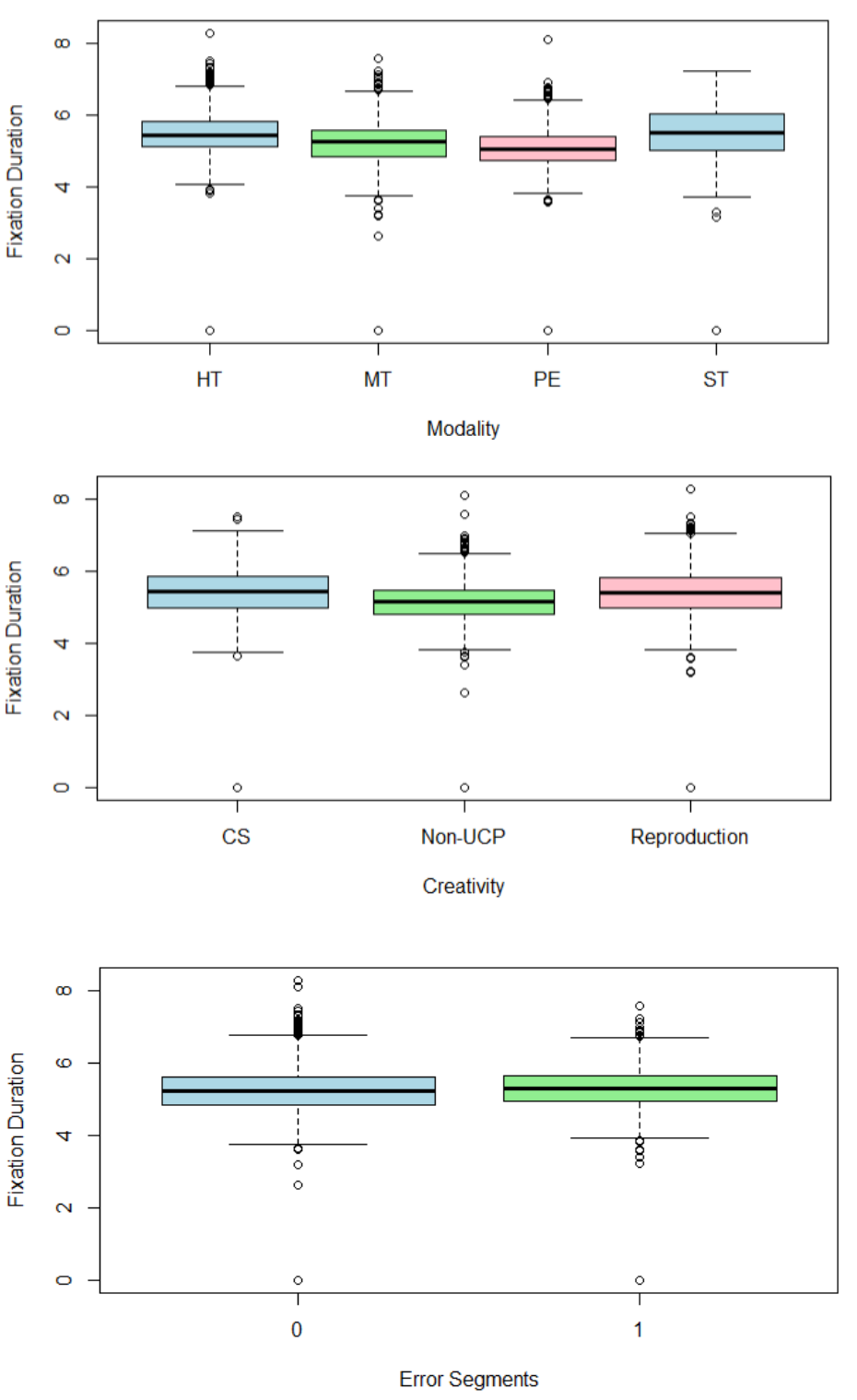}
    \caption{Box plots for the eye-tracking data (dataset II), with TFD by Modality, Creativity and Errors.}
    \label{fig:Boxplot_mod}
\end{figure}

Table \ref{table:Descriptive_res} and Figure \ref{fig:Boxplot_mod} show mean TFDs according to the variables Modality, Creativity and Errors, for dataset II (values normalised according to number of words per unit). We use TFD as this is often seen as general and overall indication of cognitive load. For Modality, ST has the highest mean duration, followed by HT, MT and then PE. This result might seem surprising: MT has the most errors (see Table \ref{table:1}) and previous studies have shown that errors in translated texts lead to an increase in cognitive load (\citealp{Kasperaviciene2020}; \citealp{stymne-etal-2012-eye}). A potential cause could be the scale of text looked at: previous studies analyse cognitive load for individual sentences with and without errors, whereas we look at the entire text. For example,  \citet{colman-etal-2022-geco}, who also looked at an entire text instead of individual sentences, also found no significant effects for modality. Furthermore, literary reading studies have shown that immersivity and engagement also increase cognitive loads in literary texts; thus, a higher quality in the literary translation--as in HT and PE--could explain a higher cognitive load and hence a higher TFD value.

\begin{table}[h!]
\scriptsize
\centering
\begin{tabular}{ l l  l  l  l  l  l }
\hline
\multirow{1}{*}{\textbf{TFD (ms.)}} &  & \textbf{n} & \textbf{Mean (SD)} & \textbf{Median} & \textbf{Min} & \textbf{Max} \\
\hline
\multirow{4}{*}{Modality} & HT & 918 & 297 (260) & 234 & 0 & 4042 \\
& MT & 896 & 219 (161) & 191 & 0 & 1974 \\
& PE & 880 & 193 (171) & 158 & 0 & 3329 \\
& ST & 924 & 311 (221) & 248 & 0 & 1386 \\
\hline
\multirow{5}{*}{Creativity} &  CS & 360 & 296 (251) & 234 & 0 & 1840 \\
& Rep. & 728 & 288 (266) & 222 & 0 & 4042 \\
& Not & 1606 & 201 (152) & 175 & 0 & 3329 \\
& UCP* & 370 & 326 (195) & 294 & 0 & 880 \\
& Not* & 554 & 302 (238) & 230 & 0 & 1386 \\
\hline
\multirow{2}{*}{Errors} & Yes & 692 & 239 (171) & 199 & 0 & 1974 \\
& No & 202 & 237 (219) & 176 & 0 & 4042\\
\hline
\end{tabular}
\caption{Overview of the eye-tracking data for TFD on each independent variable. UCP* and Not* refer to creative potential in ST rather than in translation.}
\label{table:Descriptive_res}
\end{table}

If we compare the TFD according to creativity, we find that CSs and Reproductions (UCPs) have a higher mean than non-UCP segments, indicating that UCPs have a higher cognitive load than those segments without UCPs, but CSs do not show a difference in cognitive load compared to Reproductions. In the ST, UCPs also have a higher mean TFD compared to non-UCPs--included in the table with asterisks--perhaps due to the
foregrounded elements (\citealp{Jacobs2015}; \citealp{Muller2017}). For Errors, as expected, units with errors have a higher mean TFD than those without, although this difference is minimal, as the box plot also shows. Moreover, standard deviations are high indicating high variability across the data. This is expected due to the small number of participants and the fine-grained nature of our experiment--measuring cognitive load on word-level creates much variance, even within participants.

\subsubsection{GAM analysis: main effects and interactions for TFD}

\begin{table}[h!]
\scriptsize
\centering
\begin{tabular}{ l | l  l  l  l  }
\hline
\multirow{1}{*}{\textbf{Effects}} & \textbf{Levels}& \textbf{Mean} & \textbf{SD} & \textbf{p-value} \\
\hline
\multirow{2}{*}{\makecell[l]{Inter- \\cept}} & & 1.674 & 0.0164 & N/A \\
& & & & \\
\hline
\multirow{4}{*}{\makecell[l]{Mod- \\ ality}} & HT & 0.0748 & 0.0396 & 0.059 \\
& MT & -0.0230 & 0.0396 & 0.743 \\
& PE & -0.0489 & 0.0688 & 0.477 \\
& HT (v.PE) & 0.1237 & 0.0795 & 0.120 \\
\hline
\multirow{2}{*}{\makecell[l]{Crea- \\ tivity}} &  CS & 0.0356 & 0.0084 & 2.6x10$^{-5***}$ \\
& Rep. & 0.0569 & 0.0069 & 2.5x10$^{-16***}$ \\
\hline
\multirow{1}{*}{Errors} & Yes & 0.0009 & 0.0054 & 0.867 \\
\hline
\multirow{8}{*}{\makecell[l]{Inter- \\actions \\ between \\ modal- \\ ity \\ \& \\ creativ- \\ ity}} & HT : CS & -0.0069 & 0.0140 & 0.621  \\
& MT : CS & -0.0452 & 0.0173 & 0.009$^{**}$ \\
& PE : CS & 0.0834 & 0.0334 & 0.012$^{*}$ \\
& \makecell[l]{HT (v.PE) \\ : CS} & 0.0765 & 0.0378 & 0.042$^{*}$ \\
\cline{2-5}
& HT : Rep &  0.0299 & 0.0135 & 0.027$^{*}$ \\
& MT : Rep & -0.0074& 0.0102 & 0.470 \\
& PE : Rep & 0.0447 & 0.0175 & 0.011$^{*}$ \\
& \makecell[l]{HT (v.PE) \\ : Rep} & 0.0747 & 0.0237 & 0.001$^{*}$ \\
\hline
\end{tabular}
\caption{Main effects and relevant interaction effects from the GAM model on TFD (log-transformed duration data in ms.), ***p <.001, **p <.01, *p <.05}
\label{table:GAMM}
\end{table}

The results for the GAM analysis are partially shown in Table \ref{table:GAMM}, including the main effects and relevant interactions; the full table is in Appendix \ref{sec:appendix-ET}. The only significant main effect is Creativity--both CSs and Reproductions. This indicates that there is an increase in cognitive load in UCPs overall (RQ1), although no clear difference between CSs and Reproductions (RQ3), as we saw for the descriptive statistics, too. There are no significant results for the independent variable Modality or Errors (RQ4). Effect sizes are less meaningful in this setting, as the GAM model is non-linear and uses log-transformations and a log-link; for an indication of differences between the levels of the separate independent variables; however, the mean TFD reported in Table \ref{table:Descriptive_res} illustrates this effect. 

There are also significant values in the interaction between Modality and Creativity (RQ2). MT has a negative effect in both interactions, significantly so for CS. So, although CSs have an increased cognitive load overall, this effect is lessened for readers in MT. We see the opposite for PE and HT, where the effect for CSs is increased for PE, and the effect of Reproductions is increased for PE and HT. Furthermore, comparing HT and PE directly reveals that in HT readers exert significantly more cognitive load on CSs and Reproductions than in PE. Readers thus seem to process CSs and Reproductions differently in different modalities, and this seems to indicate that HT has a higher level of cognition and attention in UCPs (CSs and Reproductions) (RQ2). There were no significant interaction effects for Errors with Modality or with Creativity (RQ4).

\subsubsection{Frequency analysis}
Lastly, we checked the effect of word frequency on our data. We correlated the word frequencies with TFD, using Spearman's correlation after checking assumptions. Results show a low negative correlation between the variables Word frequency and TFD ($\rho$ = -0.30), although significant (p < 0.005). The correlations between word frequency and the other dependent variables show similar results, see Appendix \ref{sec:appendix-ET}. Although an effect of frequency was expected, the low correlation here shows our reading measures are likely influenced by other factors than frequency alone. 

\begin{figure}[h]
    \centering
     \includegraphics[scale=0.24]{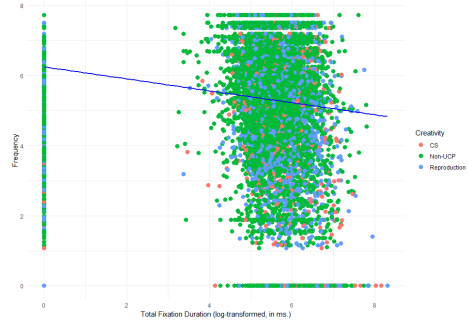}
    \caption{Scatter plot of word frequency and TFD (log-transformed). Colouring indicates Creativity annotation, showing no clear trend. }
    \label{fig:Frequency_corr}
\end{figure}

We also created a scatter plot for the correlation, with colour-coding for Creativity to see if words assigned certain Creativity codes (CS, Rep., non-UCP) tended to be more or less frequent, see Figure \ref{fig:Frequency_corr}. Colour-coding reveals no pattern, showing no clear difference between word frequencies for words belonging to CSs, Rep. and Non-UCPs. This suggests that words in UCPs (CS and Rep.) are not necessarily less frequent than words outside UCPs.

\subsection{Retrospective reading experiences}
The RTA interviews were between 10 and 25 minutes per person, with a total of 144 min. of recording. This was transcribed --translated if necessary-- resulting in 17732 words of transcription. One of the researchers coded the data in three cycles of coding. First, parallel coding was used, combining coding techniques such as emotion coding, concept coding and process coding \citep{Saldana}. The focus was on the participants' (emotional) responses, what they commented on and how. This helped to understand how readers related to the story and thus helped to interpret differences in cognitive load, and to see whether and how they reacted to elements of creativity and error. In the next cycle the initial codes were combined into 11 code groups. In the final cycle, the code groups were distributed in five main themes, see Table \ref{table:3}. Due to space limitations, a summary per theme is given, with a detailed analysis in Appendix \ref{sec:appendix-RTA}. 

\begin{table}[h!]
\small
\centering
\begin{tabular}{ p{5.5 cm}  l }
\hline
 \textbf{Theme} & \textbf{\#Codes} \\ 
\hline
Confusion came from the narrative in HT, but from language use in MT & 226\\  
 \hline
Engaging with and relating to narrative elements occurred in HT, ST \& PE  & 103 \\
 \hline
HT participants felt immersed in the story, the narrative, and the style & 82 \\
 \hline
MT participants had difficulty understanding the text due to nonsensical words phrasing & 175 \\
 \hline
PE participants were engaged in the narrative, but struggled with the style and characters at times & 106 \\
\hline

\hline
\end{tabular}
\caption{Main themes with number of codes included}
\label{table:3}
\end{table}

\textit{1. Confusion came from the narrative in HT, but from language use in MT} \\
All participants mentioned being confused (30x HT, 37x MT, 28x PE, 23x ST). However, the cause for confusion was different across the different groups. HT participants felt confused about narrative elements, rather than phrasing:  "I was a little confused by this, but that wasn't necessarily due to the words but due to the narrative" (P06\_HT). Their confusion cleared after figuring out the story more. This was similar for ST, where participants were also confused  about the narrative at first but this lessened as the narrative was revealed. However, MT participants were mostly confused about words or phrases that were translated incorrectly, mentioning multiple times "[it] didn't make sense" (P02\_MT) and that the text was "weird" (P07\_MT). They also mentioned (incorrectly translated) UCPs which caused confusion, saying in one instance: "I was completely confused (...) No idea what they did here or were intending to do." (P07\_MT). 

\textit{2. Engaging with and relating to narrative elements occurred in HT, ST \& PE participants}\\
HT, ST and PE participants mentioned feeling immersed in the narrative, relating it to their own lives multiple times. Participants mentioned that the story was engaging, with a well set up moral dilemma, making them empathise with Edward’s (the protagonist of the story) choices: "I really sympathised with the father" (P09\_PE), "I thought it was intriguing and felt bad for that man" (P08\_ST), "It was sad, but it also makes you think" (P06\_HT). 

\textit{3. HT participants felt immersed in the story, the narrative, and the style}\\
HT participants reported feeling immersed in both the narrative and the style throughout ("I really got into the story" (P03\_HT))--something that lacked for the other modalities. HT participants specifically appreciated the style ("That was brought across very well" (P06\_HT)) and repeated how much they liked the characters. Both also specifically appreciated certain translation solutions for wordplay and metaphors that were parts of UCPs.

\textit{4. MT participants had difficulty understanding the text due to nonsensical words phrasing} \\
Most salient for MT participant was their confusion regarding the language use. Both participants did not understand many phrases or details in the story, which made it difficult to follow the story along, as they tried reconstructing the story by working back from the words to “what they should have been” (P02\_MT). Both also had to laugh multiple times due to the strangeness of the MT output.  P07\_MT encapsulated this saying “It almost becomes poetical how bad it is." This also caused them to skim read later parts of the text as they gave up trying to understand the text with blatant errors; this might explain the lack of effect for errors on cognitive load in the eye-tracking: although errors (in MT) impact reader experience, the expected increase in cognitive load could be nullified by skim reading. 

\textit{5. PE participants were engaged in the narrative, but struggled with the style and characters} \\
PE participants were positive about the narrative, but they disliked the style, feeling it was sometimes used incorrectly or off-putting, hampering their overall enjoyment and immersion in the story. They found the moral dilemma interesting and liked the set-up, but also commented on "awkward" phrases (P09\_PE) or words used out of context. P05\_PE mentioned she "realised it had to be a translation, because no Dutch person would have written it like this”. They also felt character descriptions were unclear or wrong, with uncommon labels ("\textit{broeder} just confused me" (P05\_PE) and oddly used adjectives (P09\_PE). This could be related to the relatively low score for comprehension in PE (see Table \ref{table:questionnaire} and Appendix \ref{sec:appendix-RTA} for more). However, this lack of comprehension could also be due to individual differences between participants.

\section{Conclusions and further research}

We were seeking to test our methodology with four RQs that linked cognitive load and creativity. For ease of understanding we present here the questions and the findings, followed up by the limitations of the study and future avenues of research.

\textbf{RQ1: Do readers have a higher cognitive load in units of creative potential than in other regular parts of the sentence?}\\
There was a strong positive effect of UCPs on cognitive load, both in CSs and Reproductions. Readers thus pay attention to UCPs when reading, and judging from their comments in the RTAs, they also enjoy reading them. This link between engagement and cognitive load was also found in literary reading studies \citep{Torres2021}, specifically for foregrounded elements (\citealp{Jacobs2015}; \citealp{Muller2017}). 
Although the methodology employed is solid and the results novel, the creation of the datasets is quite arduous. We think that it might be better for this word-level experiments to look at datasets that have been purposely created for this, for example, by only looking at paragraphs with specific UCPs where it will be easier to explore the effect of modality.  

\textbf{RQ2: Do readers process these units differently according to the translation modality?} \\ Although we did not find an effect for Modality, we did see a positive effect in its interaction with Creativity when looking at TFD. Furthermore, the effect of Creativity was lessened in MT, while increased in both PE and HT. Comparing HT and PE against each other specifically, we see that in HT the effect of Creativity is the highest. This was reinforced in the RTAs, where HT participants were more positive about the text, especially its style, even explicitly appreciating certain translation solutions. PE participants liked the narrative, but thought the style fell flat at times. This differs from \citet{Whyatt2024}, who found increased cognitive load for low quality sentences; however, they only looked at non-literary texts, while we looked at a literary text, and literary studies have shown a positive link between the literariness of a text and fixations (\citealp{Corcoran2023}; \citealp{Fechino2020}), so that might explain the difference. Potentially, the higher overall quality of HT (as attested in the RTAs) enhances the immersive effects of UCPs.

\textbf{RQ3: Do readers process the translators' solutions differently according to the level of creativity?} \\ We did not see any significant difference in cognitive load between CSs and Reproductions. Analysing word frequency also showed only a low negative correlation between word frequency and TFD, showing that other factors influenced TFD. There was also no clustering of CSs, Reproductions. or non-UCPs, indicating that words belonging to CSs or Reproductions are not less frequent than words that do not belong to UCPs. Again here, we think that, methodologically, experiments of this type would benefit for more focused studies at paragraph-level with chosen UCPs.

\textbf{RQ4: Do errors in a segment increase the cognitive load of the reader?} \\ As expected, we see that units with errors have a higher mean TFD than those without, although only slightly. However, we do not see any significant effect on units with or without errors. This differs from previous studies on errors, as \citet{Kasperaviciene2020} and \citet{stymne-etal-2012-eye} found increased TFD and FC on sections with errors. Triangulating the data with the interviews, however, allowed us to hypothesise that this is due to increased skim reading in MT especially, the modality with most errors. When these readers saw too many errors, they tended to skim certain sections more. RTAs showed that errors in MT and PE influenced reading experience --creating a lack of understanding in MT and a dislike of style in PE-- but this was not mirrored in the exerted cognitive load. 

This study shows both the methodological advantages of analysing reader reception through cognitive load on a word-level, and the importance of creativity for reader reception across modalities. Intuitively, we expect differences in reading experiences across modalities, given errors and creativity scores. In previous studies, PE and HT scored similarly and we initially also found no main effect on cognitive load between the two; however, detailed analysis shows that creativity makes a difference; not only that, but the difference creativity makes is increased in HT. In other words, the effect of creativity is lessened in PE compared to HT. The retelling of the participants' experience through the RTAs also showed clear differences between the modalities, a crucial aspect of this methodology. 

Using MT-mediated texts in literary translation thus has an impact on readers. More research into the causes of these effects is needed to inform translation technologies, translators and the industry.

We are aware of the limitations of this pilot regarding number of participants and language pairs. However, the methodology gives us a window through which we can explore the way readers deal with creativity while reading. We learnt from this experiment that document-level analysis might not be the best match to answer our RQs, therefore our next experiment, focuses on paragraphs with selected UCPs in different modalities. This will include more participants, different genres, and LLMs, to further explore this relation between creativity and reader experience.
\section*{Acknowledgments}
We would like to thank all our participants. We would also like to thank Andreas van Cranenburgh for his assistance with the frequency analysis.
\section*{Funding}
The INCREC project has received funding from the European Union’s Horizon Europe research and innovation programme under ERC Consolidator Grant n. 101086819. 

\bibliography{anthology, mtsummit25}

\newpage
\appendix

\section{Sustainability Statement: Energy costs and CO2 Emission Related to Experiments}
Experiments in this paper made use of already existing contents (the raw MT output had been created for the previous experiment by \citet{Guerberof2024} and no additional models were trained, optimised or used. Therefore, there were no energy costs or carbon dioxide emissions for computational efforts related to the creation of this paper \citep{lacoste2019quantifying}.
                       
\section{Appendix: Questionnaire}   
\label{sec:appendix-Q}
The questionnaire was created in English and then translated into Dutch. As explained above, the questionnaire has a pre-task and a post-task part. The pre-task part focuses on demographics and reading habits. The demographics included questions on gender, age, education, employment and native language. The questions on reading habits asked about how often participants read, how much they enjoy reading, in which ways they read (physical book, e-book, audiobook, tablet, laptop, etc.), in which language they read (percentage-based), which genres they prefer, how often they read in Dutch and how long they read for typically.

The post-task part of the questionnaire consisted out of four sections. The first section was comprehension and was related to details of the story and were multiple-choice. As we did not discuss the story in detail in the article, we decided not to include all questions (and potential answers) here--they can however be found on \url{https://github.com/KyoGerrits/To-MT-or-not-to-MT}. 

The other three section were scored on a 7-point Likert scale. For \textbf{narrative engagement} the questions were:
\begin{enumerate}
    \item At times, I struggled to understand what was happening in the story
    \item My understanding of the character is unclear
    \item  I had a hard time recognizing the thread of the story
    \item My mind wandered while reading the text
    \item While reading, I found myself thinking about other things
    \item I had a hard time keeping my mind on the text
    \item While reading, my body was in the room, but my mind was inside the world created by the story
    \item The text created a new world, and then that world suddenly disappeared when the story ended
\item At times when reading, the story world was closer to me than the real world
\item During the story, I felt sad when a main character suffered in some way.
\item The story affected me emotionally.
\item I felt sorry for some of the characters
\item While reading the story I had a clear image of what the main character looked like.
\item While reading the story I could envision the situations described
\item I could imagine what the setting of the story looked like.
\end{enumerate}

For \textbf{enjoyment}:
\begin{enumerate}
    \item Did you enjoy the text?
    \item How likely is it that you would recommend the text to a friend?
    \item Would you consider this text high literature?
\end{enumerate}

For \textbf{translation reception}:
\begin{enumerate}
    \item The text was easy to understand
    \item The text was well-written
    \item I encountered words, sentences or paragraphs that were difficult to understand (including a box to write down which ones)
    \item I encountered words, sentences or paragraphs that I found very beautiful (including a box to write down which ones)
    \item I noticed I was reading a translation (including a box to indicate how people noticed)
    \item What did you think of the translation?
    \item Would you like to read a text by the same author and translator?
    \item Would you like to read a text by the same author, but by a different translator?
    \item Would you like to read a text by a different author, but the same translator?
\end{enumerate}

In \citet{Guerberof2024}'s study, the
Cronbach’s alpha reliability coefficient ($\alpha$) was 0.85 for narrative engagement, 0.87 for enjoyment and 0.79 for translation reception. These are good scores for reliability and shows the reliability of the scales. For completeness' sake, we also calculated Cronbach's alpha with our data. We had scores respectively of 0.847, 0.813, and 0.915, of which the first two are considered good and the final one excellent.

\section{Appendix: Eye-tracking statistics}
\label{sec:appendix-ET}
\subsection{Overview eye-tracking results per word (dataset I)}
This section includes the overview of the eye-tracking data for all the dependent variables (TFD, FPT, RP, FC \& RC) according to our independent variables (Modality, Creativity and Error) in dataset I, that is, the dataset per word instead of per unit, dataset II, as was used in the main analysis.

Tables \ref{table:TFDw_overview}, \ref{table:FPTw_overview}, \ref{table:RPw_overview}, \ref{table:FCw_overview}, and \ref{table:RCw_overview} show the descriptive values for each of our dependent variables (TFD, FPT, RP, FC, and RC) per word. Comparing the table with the descriptive results for the dependent variables per unit (see Table \ref{table:Descriptive_res} for TFD and the Tables CHECK below for the other dependent variables per unit), we see similar results as we saw in the descriptive results per unit (dataset II).

\begin{table}[h!]
\scriptsize
\centering
\begin{tabular}{ l  l l  c  l  l  l }
\hline
\multirow{1}{*}{\textbf{IVs}} & \textbf{Cat.}  & \textbf{n} & \textbf{Mean (SD)} & \textbf{Med.} & \textbf{Min} & \textbf{Max} \\
\hline
\multirow{4}{*}{Modality} & HT & 5164 &  257 (275) & 205 & 0 & 4042 \\
& MT & 5204 & 192 (199) &177 & 0 & 2266\\
& PE & 5160 &  177 (215)&  156	& 0 &	3034 \\
& ST & 5096 &  316 (378) & 210 & 0 & 3629 \\
\hline
\multirow{3}{*}{Creativity} & CS & 1350 & 259 (291) & 204 & 	0 &	3484 \\
& Rep. & 2948 & 243 (264) & 199 & 	0 &	4042 \\
& Not & 11230 & 193 (216) & 171 &	0 &	2560 \\
\hline
\multirow{2}{*}{Errors} & Yes & 1704 & 	243  (249) & 198 & 	0	& 3034 \\
& No & 13824 & 205 (232) & 176& 	0 &	4042\\
\hline
\end{tabular}
\caption{Overview of the eye-tracking data for TFD (in ms.) on each independent variable, per word (dataset I)}
\label{table:TFDw_overview}
\end{table}

\begin{table}[h!]
\scriptsize
\centering
\begin{tabular}{ l  l l  c  l  l  l }
\hline
\multirow{1}{*}{\textbf{IVs}} & \textbf{Cat.}  & \textbf{n} & \textbf{Mean (SD)} & \textbf{Med.} & \textbf{Min} & \textbf{Max} \\
\hline
\multirow{4}{*}{Modality} & HT & 5164 & 158 (124) & 169 & 	0 & 	995 \\
& MT & 5204 & 136 (115) & 155 & 	0 &	1067 \\
& PE & 5160 &  124 (121) & 138 & 	0 &	981 \\
& ST & 5096 &  152 (127) & 158 & 	0 &	1129 \\
\hline
\multirow{3}{*}{Creativity} & CS & 1350 & 156 (122) & 168 & 	0 & 	803 \\
& Rep. & 2948 & 154 (124) &  166 & 	0 &	981 \\
& Not & 11230 & 134 (119) & 	150 & 	0 &	1067 \\
\hline
\multirow{2}{*}{Errors} & Yes & 1704 & 	152 (118) &	165 & 	0	& 803 \\
& No & 13824 & 138 (121) & 153 & 0 & 1067\\
\end{tabular}
\caption{Overview of the eye-tracking data for FPT (in ms.) on each independent variable, per word (dataset I)}
\label{table:FPTw_overview}
\end{table}

\begin{table}[h!]
\scriptsize
\centering
\begin{tabular}{ l  l l  c  l  l  l }
\hline
\multirow{1}{*}{\textbf{IVs}} & \textbf{Cat.}  & \textbf{n} & \textbf{Mean (SD)} & \textbf{Med.} & \textbf{Min} & \textbf{Max} \\
\hline
\multirow{4}{*}{Modality} & HT & 5164 & 279 (506)& 202 & 0 & 	20055 \\
& MT & 5204 & 205 (351) & 173 & 0	& 8806 \\
& PE & 5160 &  214 (362) & 161 & 	0 & 	7335 \\
& ST & 5096 &  289 (535) & 191 & 	0	& 16107 \\
\hline
\multirow{3}{*}{Creativity} & CS & 1350 & 286 (446) & 	200 & 	0 &	60313 \\
& Rep. & 2948 & 264 (400) & 	197 & 	0 &	8330 \\
& Not & 11230 & 277 (552) & 185 & 	0 &	16107\\
\hline
\multirow{2}{*}{Errors} & Yes & 1704 & 	152 (118) &	165 & 	0	& 803 \\
& No & 13824 & 228 (415) & 176 & 0	 & 20005\\
\hline
\end{tabular}
\caption{Overview of the eye-tracking data for RP (in ms.) on each independent variable, per word (dataset I)}
\label{table:RPw_overview}
\end{table}

\begin{table}[h!]
\scriptsize
\centering
\begin{tabular}{ l  l l  c  l  l  l }
\hline
\multirow{1}{*}{\textbf{IVs}} & \textbf{Cat.}  & \textbf{n} & \textbf{Mean (SD)} & \textbf{Med.} & \textbf{Min} & \textbf{Max} \\
\hline
\multirow{4}{*}{Modality} & HT & 5164 & 1.201 (1.161) & 1 & 0 & 13 \\
& MT & 5204 & 0.962 (0.945) & 1 & 0 & 8 \\
& PE & 5160 & 0.865 (0.956) & 1 & 0 &	15 \\
& ST & 5096 &  1.540 (1.682) & 1 & 0 &16 \\
\hline
\multirow{3}{*}{Creativity} & CS & 1350 & 1.208 (1.240) & 1 & 0 &	13 \\
& Rep. & 2948 & 1.148 (1.127) & 1 &	0 &	15 \\
& Not & 11230 &  0.950 (0.975) & 1 & 0 & 10\\
\hline
\multirow{2}{*}{Errors} & Yes & 1704 & 	1.173 (1.113) & 1 & 0 &	11 \\
& No & 13824 & 0.990 (1.024) & 	1 &	0 &	15\\
\hline
\end{tabular}
\caption{Overview of the eye-tracking data for FC on each independent variable, per word (dataset I)}
\label{table:FCw_overview}
\end{table}

\begin{table}[h!]
\scriptsize
\centering
\begin{tabular}{ l  l l  c  l  l  l }
\hline
\multirow{1}{*}{\textbf{IVs}} & \textbf{Cat.}  & \textbf{n} & \textbf{Mean (SD)} & \textbf{Med.} & \textbf{Min} & \textbf{Max} \\
\hline
\multirow{4}{*}{Modality} & HT & 5164 & 0.204 (0.475) & 0 & 0 &	5 \\
& MT & 5204 & 0.142 (0.574) & 0 & 0 &29\\
& PE & 5160 & 0.151 (0.425) & 0 & 	0 &	10 \\
& ST & 5096 & 0.165 (0.451) &0 & 0 &	6 \\
\hline
\multirow{3}{*}{Creativity} & CS & 1350 & 0.221 (0.554) & 0 & 0 &	10 \\
& Rep. & 2948 & 0.184 (0.700) & 0 &	0 &	29 \\
& Not & 11230 &  0.154 (0.418) & 0	 &0 &	6\\
\hline
\multirow{2}{*}{Errors} & Yes & 1704 & 	0.218 (0.880) & 0 &0 &29 \\
& No & 13824 & 0.159 (0.426) & 0 & 0 & 6\\
\hline
\end{tabular}
\caption{Overview of the eye-tracking data for RC on each independent variable, per word (dataset I)}
\label{table:RCw_overview}
\end{table}

For TFD, we see higher mean results here, although median scores overlap considerably. Furthermore, we see higher SDs here as well. This makes sense as all individual words are included in this dataset (dataset I) and dataset II is normalised per word over units, reducing variance. However, the trends remain the same here. For FPT, HT has a higher mean FPT than ST here, and the differences are less pronounced than in dataset II. This is similar to FC and RC in dataset I, where we see low measures overall and that the mean of HT is higher than that of ST. For RP, the high SDs across all conditions stand out. This could be caused by relatively long regressions if a word was not understood or new, while other words were looked at much easier and quicker. In terms of mean and median values, we still see comparable results as in the other dependent variables and as in the  dataset II. 

\begin{figure}[h]
    \centering
    \includegraphics[scale=0.35]{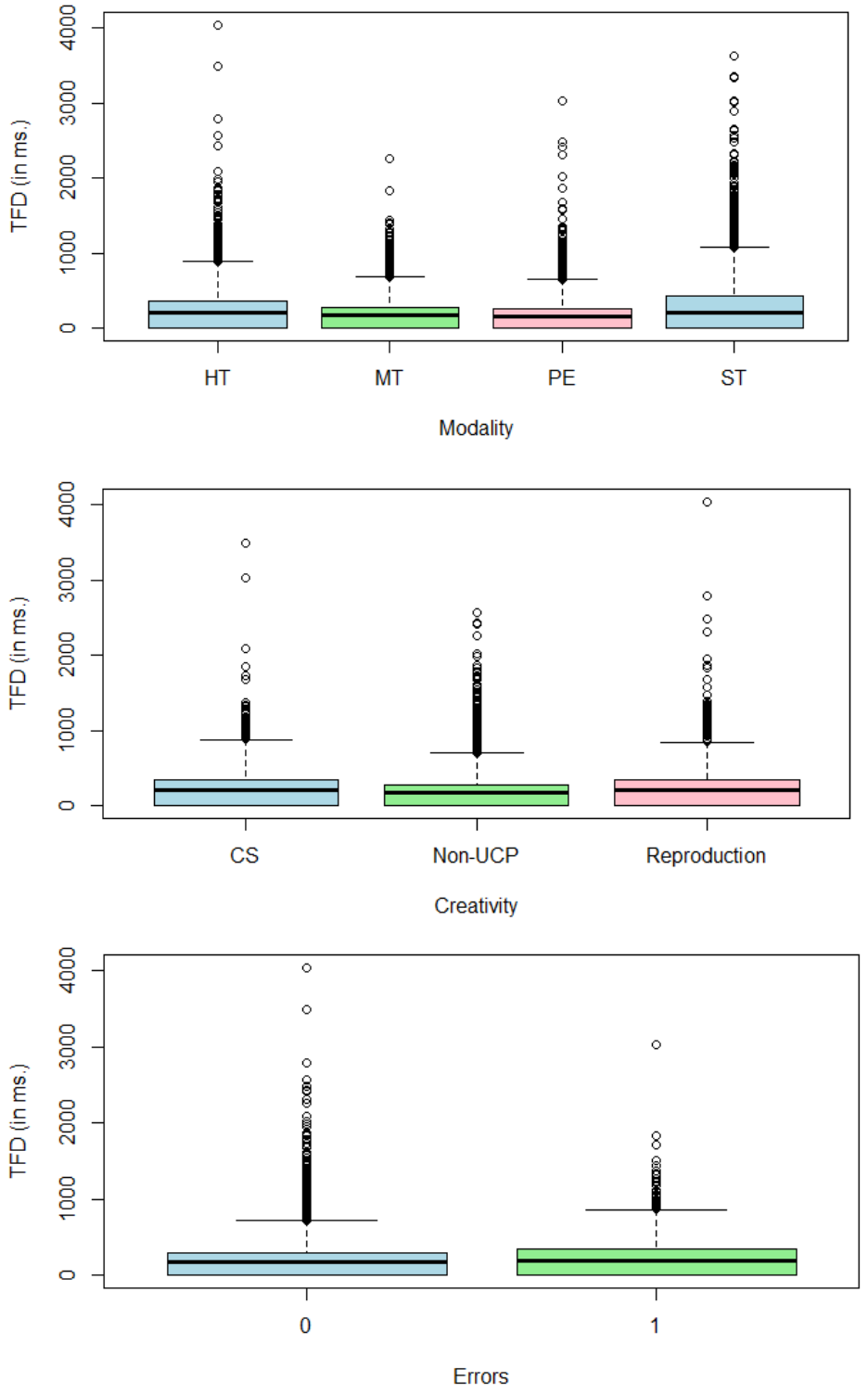}
    \caption{Box plots of TFD (in ms.) for all independent variables (Modality, Creativity and Errors), for dataset I (word-level).}
    \label{fig:boxplots_words}
\end{figure}

We also include the box plots for TFD for the independent variables (Modality, Creativity and Errors) for all units per word. These are shown in Figure \ref{fig:boxplots_words}. These figures show similar results as Figure \ref{fig:Boxplot_mod}, with HT and ST showing slightly higher TFD values compared to MT and PE. CS and Rep also have slightly higher values than Non-UCPs. Finally, the units that have a presence of errors has narrowly higher TFD scores than those without errors. The box plots for the other dependent variables show a similar trend, with little difference between the conditions. 

\subsection{Overview eye-tracking data for FPT, RP, FC, RC}
This section shows the descriptive values of the eye-tracking data for our other dependent variables (FPT, RP, FC, RC) for database II. The results for TFD are in the main body, see Table \ref{table:Descriptive_res}.

\begin{table}[h!]
\scriptsize
\centering
\begin{tabular}{ l  l l  l  l  l  l }
\hline
\multirow{1}{*}{\textbf{IV}} & \textbf{Cat.}  & \textbf{n} & \textbf{Mean (SD)} & \textbf{Med.} & \textbf{Min} & \textbf{Max} \\
\hline
\multirow{4}{*}{Modality} & HT & 918 & 164 (71) & 159 & 0 & 644	 \\
& MT & 896 & 145 (64) & 139 & 0 & 484\\
& PE & 880 &  130 (81) & 122 & 0 & 1644 \\
& ST & 924 &  158 (79) & 150 & 0 &	584 \\
\hline
\multirow{3}{*}{Creativity} & CS & 359 & 165 (79) & 161 & 0 & 674 \\
& Rep. & 728& 158 (72) & 152 & 0 & 644 \\
& Not & 1607 & 137 (72) & 131 & 0 & 1644 \\
\hline
\multirow{2}{*}{Errors} & Yes & 692 & 	151 (67) & 142 & 0 &	674 \\
& No & 2002 & 145 (76) & 138 & 0 & 1644\\
\hline
\end{tabular}
\caption{Overview of the eye-tracking data for FPT (in ms., normalised for words per unit (dataset II)) on each independent variable.}
\label{table:FPT_overview}
\end{table}

\begin{table}[h!]
\scriptsize
\centering
\begin{tabular}{ l  l l  l  l  l  l }
\hline
\multirow{1}{*}{\textbf{IV}} & \textbf{Cat.}  & \textbf{n} & \textbf{Mean (SD)} & \textbf{Med.} & \textbf{Min} & \textbf{Max} \\
\hline
\multirow{4}{*}{Modality} & HT & 918 & 319 (346) & 241 & 0 & 4042 \\
& MT & 896 & 250 (261) & 199 & 	0 &	4322\\
& PE & 880 & 232 (246) & 171 & 0 & 3663 \\
& ST & 924 & 429 (536)  & 258  & 0	& 3921 \\
\hline
\multirow{3}{*}{Creativity} & CS & 359 & 331 (329) & 247&  0 &	2774 \\
& Rep. & 728& 317 (374) & 228 & 0 & 4322 \\
& Not & 1607 & 231 (226) & 189 & 0 & 3775 \\
\hline
\multirow{2}{*}{Errors} & Yes & 692 & 	151 (67) & 142 & 0 &	674 \\
& No & 2002 & 264 (278) & 202 & 0	 & 4042\\
\hline
\end{tabular}
\caption{Overview of the eye-tracking data for RP (in ms., normalised for words per unit (dataset II)) on each independent variable.}
\label{table:RP_overview}
\end{table}

\begin{table}[h!]
\scriptsize
\centering
\begin{tabular}{ l  l l  l  l  l  l }
\hline
\multirow{1}{*}{\textbf{IV}} & \textbf{Cat.}  & \textbf{n} & \textbf{Mean (SD)} & \textbf{Med.} & \textbf{Min} & \textbf{Max} \\
\hline
\multirow{4}{*}{Modality} & HT & 918 & 1.363 (1.012) & 1 & 0 & 10 \\
& MT & 896 & 1.085 (0.748) & 1 &	0	 & 9\\
& PE & 880 & 1.002 (0.812) & 0.88 &	0 &	14 \\
& ST & 924 & 1.604 (1.173) & 1.25 &	0 & 9 \\
\hline
\multirow{3}{*}{Creativity} & CS & 359 & 1.389 (1.067) & 1 &	0 &	7.3 \\
& Rep. & 728& 1.383 (1.074) & 1 & 0 & 10 \\
& Not & 1607 & 0.995 (0.678) & 	1	 & 0	& 14 \\
\hline
\multirow{2}{*}{Errors} & Yes & 692 & 	1.176 (0.796) & 	1	 & 0 &	9 \\
& No & 2002 & 1.144 (0.907) & 	1	 & 0	 & 14\\
\hline
\end{tabular}
\caption{Overview of the eye-tracking data for FC (normalised for words per unit (dataset II)) on each independent variable.}
\label{table:FC_overview}
\end{table}

\begin{table}[h!]
\scriptsize
\centering
\begin{tabular}{ l  l l  l  l  l  l }
\hline
\multirow{1}{*}{\textbf{IV}} & \textbf{Cat.}  & \textbf{n} & \textbf{Mean (SD)} & \textbf{Med.} & \textbf{Min} & \textbf{Max} \\
\hline
\multirow{4}{*}{Modality} & HT & 918 & 0.177 (0.305)& 0 & 0 & 3 \\
& MT & 896 & 0.152 (0.312) & 0 & 0 & 4\\
& PE & 880 & 0.179 (0.284)	& 0.10 & 0	& 4 \\
& ST & 924 & 0.459 (0.559) & 0.33&	0 &	7 \\
\hline
\multirow{3}{*}{Creativity} & CS & 359 & 0.198 (0.298) & 0 & 0 & 1.67  \\
& Rep. & 728& 0.198 (0.299) & 0 & 0 & 1.67  \\
& Not & 1607 & 0.151 (0.251) & 0.06 & 0 & 4 \\
\hline
\multirow{2}{*}{Errors} & Yes & 692 & 	0.171 (0.323) & 0.04 &0 & 	4 \\
& No & 2002 & 0.168 (0.292) & 0 & 0 & 4\\
\hline
\end{tabular}
\caption{Overview of the eye-tracking data for RC (normalised for words per unit (dataset II)) on each independent variable.}
\label{table:RC_overview}
\end{table}

Comparing the descriptive values of FPT, RP, FC, and RC in Tables \ref{table:FPT_overview}, \ref{table:RP_overview}, \ref{table:FC_overview} and \ref{table:RC_overview} with those in Table \ref{table:Descriptive_res} for TFD, we see similar results across the board with some small difference. For FPT, we see that mean score for HT is higher than the mean score of ST. For RP, on the other hand, the difference between ST and the other modalities is higher. For FC, the only difference is that the differences between the values are smaller than they were for TFD. For RC, what stands out are the low scores (almost all between 0.151 and 0.198, with 5 out of 9 categories with a median score of zero).

\subsection{GAM analysis}
\begin{table}[h!]
\scriptsize
\centering
\begin{tabular}{ l | l  l  l  l  }
\hline
\multirow{1}{*}{\textbf{Effects}} & \textbf{Levels}& \textbf{Mean} & \textbf{SD} & \textbf{p-value} \\
\hline
\multirow{2}{*}{\makecell[l]{Inter- \\cept}} & & 1.674 & 0.0164 & N/A \\
& & & & \\
\hline
\multirow{4}{*}{\makecell[l]{Mod- \\ ality}} & HT & 0.0748 & 0.0396 & 0.059 \\
& MT & -0.0230 & 0.0396 & 0.743 \\
& PE & -0.0489 & 0.0688 & 0.477 \\
& HT (v.PE) & 0.1237 & 0.0795 & 0.120 \\
\hline
\multirow{2}{*}{\makecell[l]{Crea- \\ tivity}} &  CS & 0.0356 & 0.0084 & 2.6x10$^{-5***}$ \\
& Rep. & 0.0569 & 0.0069 & 2.5x10$^{-16***}$ \\
\hline
\multirow{1}{*}{Errors} & Yes & 0.0009 & 0.0054 & 0.867 \\
\hline
\multirow{8}{*}{\makecell[l]{Inter- \\actions \\ between \\ modal- \\ ity \\ \& \\ creativ- \\ ity}} & HT : CS & -0.0069 & 0.0140 & 0.621  \\
& MT : CS & -0.0452 & 0.0173 & 0.009$^{**}$ \\
& PE : CS & 0.0834 & 0.0334 & 0.012$^{*}$ \\
& \makecell[l]{HT (v.PE) \\ : CS} & 0.0765 & 0.0378 & 0.042$^{*}$ \\
\cline{2-5}
& HT : Rep &  0.0299 & 0.0135 & 0.027$^{*}$ \\
& MT : Rep & -0.0074& 0.0102 & 0.470 \\
& PE : Rep & 0.0447 & 0.0175 & 0.011$^{*}$ \\
& \makecell[l]{HT (v.PE) \\ : Rep} & 0.0747 & 0.0237 & 0.001$^{*}$ \\
\hline
\multirow{4}{*}{\makecell[l]{Inter- \\actions \\ between \\ Mod. \& \\ Errors}} & HT : Error & -0.0064 &  0.0125 &  0.610 \\
&MT : Error & 0.0013 &  0.0124 &  0.914 \\
&PE : Error & -0.0091 &  0.0232 &  0.696 \\
& \makecell[l]{HT (v.PE) \\ : Error} &   -0.0155 &  0.0278 &  0.579 \\
\hline
\multirow{3}{*}{\makecell[l]{Interact. \\ Crea. \& \\ Errors}} & CS : Errors & -0.0209 &   0.0135 &  0.122 \\
& Rep : Errors &  -0.0104 &   0.0103 & 0.308 \\
& & & &  \\
\hline
\multirow{4}{*}{\makecell[l]{Three way \\ inter- \\ actions \\ between \\ Modality, \\ Creativity \\ \& Errors}} & HT : CS : Error & -0.0314 &  0.0294 & 0.285 \\
& MT : CS : Error &  0.0047 &  0.0332 & 0.888 \\
& PE : CS : Error & -0.0408 &  0.0629 & 0.517 \\
& \makecell[l]{HT (v.PE) \\ : CS : Error} & -0.0722 &  0.0724 & 0.318 \\
\cline{2-5}
& HT : Rep : Error & 0.0667 & 0.0281 & 0.018$^{*}$ \\
& MT : Rep : Error & 0.0183 & 0.0215 & 0.394 \\
& PE : Rep : Error & 0.0301 & 0.0370 & 0.415 \\
& \makecell[l]{HT (v.PE) \\ Rep : Error} & 0.0969 & 0.0497 &  0.051 \\
\hline
\end{tabular}
\caption{All main effects and interaction effects from the GAM model on TFD (log-transformed duration data in ms.), ***p <.001, **p <.01, *p <.05}
\label{table:GAMM_entire}
\end{table}

Table \ref{table:GAMM_entire} shows all results from the GAM analysis (whereas Table \ref{table:GAMM} in the main body only showed partial results). This includes the two-way interactions with Errors (not significant) and all three-way interactions (not relevant for the RQs).

The results for the two-way interactions with Errors are not significant, although the directions are not surprising. The negative effect for HT, PE and HT compared specifically to PE reveal that generally participants spent less cognitive load on errors in these modalities, which fit our intuition that errors in MT require a higher cognitive load than errors in the other modalities; however, this value is not significant.

\subsubsection{Analysis of segments without fixations}
\begin{table}[h!]
\centering
\small
\begin{tabular}{ l  l  l  l }
\hline
\multirow{1}{*}{\textbf{IV}} & \textbf{Cat.} & \textbf{\# of zeros} & \textbf{\% of zeros} \\
\hline
\multirow{3}{*}{Mod.} & HT & 25 & 2.7\% \\
& MT  & 17 &  1.9\% \\
& PE & 29 & 3.3\% \\
\hline
\multirow{3}{*}{Crea.} &  CS & 18 & 5\% \\
& Rep. & 12 & 1.6\% \\
& Not & 41 & 2.6\% \\
\hline
\multirow{2}{*}{Err.} & Yes & 4 & 0.6\% \\
& No & 67 & 3.3\%  \\
\hline
\multirow{1}{*}{\textbf{Total}} & & \textbf{71} & \textbf{2.6\%} \\
\hline
\end{tabular}
\caption{Frequency tables for segments without fixations (compared to the total number of segments) for each IV.}
\label{table:Zeros}
\end{table}

To analyse the segments with no fixations--as complement to the GAM-analysis--we created frequency tables for these segments for each independent variable, as seen in Table \ref{table:Zeros}. We conducted Chi-Square Goodness-of-Fit Tests for each independent variable, but there were no significant values for Modality, Creativity or Errors. There is thus no significant effect of either Modality, Creativity or Errors when participants skipped words.
 
\subsection{Non-parametric tests}
As the dependent variables FPT, RP, FC and RC did not meet assumptions for a GAM analysis, we conducted non-parametric tests for these variables across our independent variables Modality, Creativity and Errors. Non-parametric tests only work on aggregated results--one measurement per participant, or per condition when handling repeated measures--so we calculated the means per participants and per category for each of the variables, see Table \ref{table:non-par} for descriptive data for FPT, RP, FC and RC.

\begin{table}[h!]
\scriptsize
\centering
\begin{tabular}{ l |  l  l  l  l  l  l  l  }
\hline
\multirow{1}{*}{\textbf{Part.}} & \textbf{IVs} & \textbf{Lev.} & \textbf{n} & \textbf{FPT} & \textbf{RP} & \textbf{FC} & \textbf{RC} \\
\hline
\multirow{6}{*}{2B} & Mod. & MT & 2602 & 115.80 & 133.75 & 0.793 & 0.279 \\
\cline{2-8}
& & CS  & 87 & 99.07 & 100.73 & 0.558 & 0.135\\
& Crea. & Rep & 635 & 128.27 & 157.18 & 0.886 & 0.828 \\
& & Not & 1880 & 112.36 & 127.37 & 0.772 & 0.305 \\
\cline{2-8}
& Err. & Yes & 411 & 131.20 & 162.18 & 0.988 & 1.258 \\
& & No & 2191 & 112.90 & 128.42 & 0.765 & 0.474 \\
\hline
\multirow{6}{*}{3C} & Mod. & HT &2582 & 169.46 & 247.67 & 1.316& 0.155 \\
\cline{2-8}
& & CS & 312 & 178.17 & 295.24 & 1.590 & 0.247 \\
& Crea & Rep & 422 & 187.74 & 289.18 & 1.498 & 0.161 \\
&& Not & 1848 & 163.81 & 230.16 & 1.228 & 0.138 \\
\cline{2-8}
& Err & Yes & 166 & 178.07 & 268.87 & 1.470 & 0.181 \\
&& No & 2416 & 168.87 & 246.22 & 1.305 & 0.153 \\
\hline
\multirow{1}{*}{4D} & Mod. & ST & 2548 & 183.62 & 317.37 & 2.115 & 0.062 \\
\hline
\multirow{6}{*}{5A} & Mod. & PE & 2580 & 115.87 & 184.76 & 0.688 & 0.113 \\
\cline{2-8}
& & CS & 276 & 140.42 & 241.39 & 0.819 & 0.143 \\
& Crea & Rep &  417 & 128.35 & 217.87 & 0.803 & 0.100 \\
& & Not & 1887 & 163.81 & 230.16 & 1.228 & 0.138 \\
\cline{2-8}
& Err. & Yes & 275 & 138.84 & 247.28 & 0.864 & 0.148 \\
& & No & 2305 & 113.63 & 177.36 & 0.667 & 0.109 \\
\hline
\multirow{6}{*}{6C} & Mod. & HT & 2582 & 146.80 & 310.73 & 1.086 & 0.254 \\
\cline{2-8}
&& CS & 312 & 169.23 & 413.32 & 1.385 & 0.304 \\
& Crea & Rep & 422 & 168.05 & 360.88 & 1.268 & 0.296 \\
&& Not & 1848 & 138.16 & 281.96 & 0.995 & 0.236 \\
\cline{2-8}
& Err. & Yes &166 &150.54 & 300.17 & 1.102 & 	0.211 \\
& & No & 2416 & 146.54 & 311.45 & 1.085 & 0.257 \\
\hline
\multirow{6}{*}{7B} & Mod. & MT & 2602 & 156.46 &	277.87 & 1.133 & 0.169 \\
\cline{2-8}
& & CS & 87 & 167.50 &	220.55 & 1.081 & 0.163 \\
& Crea & Rep & 635 & 165.85 & 322.83 & 1.294 & 0.200 \\
& & Not & 1880 & 152.801 & 265.42 & 1.081 & 0.159 \\
\cline{2-8}
& Err & Yes & 411 & 168.44 & 368.23 & 1.457 & 0.257 \\
& & No & 2191 & 154.24 & 261.16 & 1.073 & 0.153 \\
\hline
\multirow{1}{*}{8D} & Mod. & ST & 2548 & 123.18 &	261.98 & 1.004 & 0.261 \\
\hline
\multirow{6}{*}{9A} & Mod. & PE & 2580 & 131.66 & 	241.51 & 1.036 & 0.187 \\
\cline{2-8}
&& CS & 276 & 143.29 & 254.11 & 1.192 & 0.246 \\
& Crea & Rep & 417 & 150.37 & 262.95 & 1.189 & 0.197 \\
&& Not & 1887 & 125.83 & 234.94 & 0.979 & 0.176 \\
\cline{2-8}
& Err & Yes & 275 & 158.68 & 285.09 & 1.193 & 0.255 \\
&& No & 2305 & 128.44 & 236.32 & 1.017 & 0.179 \\
\hline
\end{tabular}
\caption{Aggregated means per participant and per category for each of the variables, for the non-parametric tests. Including number of observations per category, mean FPT, mean RP, mean FC and mean RC.}
\label{table:non-par}
\end{table}

For the independent variable Modality and the dependent variables FPT, RP, FC and RC, we conducted a series of Kruskal-Wallis H tests. However, none of the results were significant. This was somewhat surprising given the differences between the modalities observed earlier for TFD. For the independent variable Creativity and the dependent variables FPT, RP, FC and RC, we conducted Friedman's Tests (RQ2). We found significant results for the dependent variables FPT ($\textit{X}^2$(2) = 6.3, p = 0.042) and FC ($\textit{X}^2$(2) = 6.3, p = 0.042), but not for the others. Post-hoc comparisons, using Wilcoxon Rank Sum tests with Holm-Bonferroni correction, did not yield significant results between levels of Creativity. We wanted to analyse our independent variable Creativity further, to see whether there was any difference between UCPs (CS and Rep.), comparing CS to Reproductions specifically (and leaving out the non-UCPs). We conducted a series of Mann Whitney U tests to compare data for TFD, FPT, RP, FC and RC for CS and Reproductions, but none of these tests were significant. The analyses show that creativity overall had an effect on our participants' cognitive load, as we had also seen in the GAM analysis, further supporting a positive answer to our second research question that readers have higher cognitive load in UCP than other units.

To look at the effect of Errors, we conducted a series of Friedman's Test for the independent variable Errors for our dependent variables  FPT, RP, FC and RC, but none of these were significant. So, although errors increase reading time in general, this is not significant as the GAM model also showed. We also checked for the effect of severity (none, minor and major) and type of error. Only the first was significant, specifically for FPT ($\textit{X}^2$(2) = 6.3, p = 0.042) and FC ($\textit{X}^2$(2) = 6.3, p = 0.042); however, here too, post-hoc Wilcoxon rank sum tests with Holm-Bonferroni correction did not reach significance.

This supports the results from the GAM analysis on TFD, with a significant result for Creativity but not for Modality and Errors. There thus seems to be increased cognitive load for UCPs (CS \& Reproductions) (RQ1), but not between CS and Reproductions (RQ3) nor for errors (RQ4).

\subsection{Frequency analyses}
We include here the frequency analyses for FPT and RP. FC and RC are not included as these are count data and do not meet the assumptions for Pearson's or Spearman's correlation. 

\begin{figure}[h]
    \centering
    \includegraphics[scale=0.4]{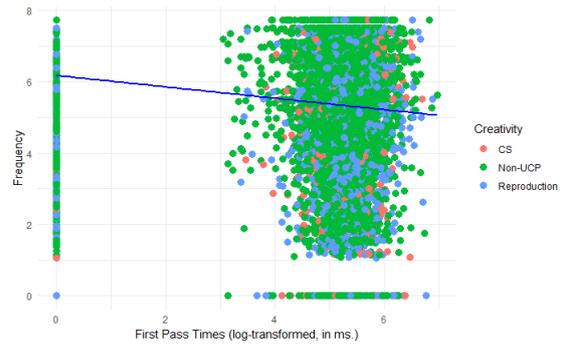}
    \caption{Scatter plot of word frequency and FPT (log-transformed). Colour indicates Creativity annotation, showing no clear trend.}
    \label{fig:SP_FPT}
\end{figure}

\begin{figure}[h]
    \centering
    \includegraphics[scale=0.4]{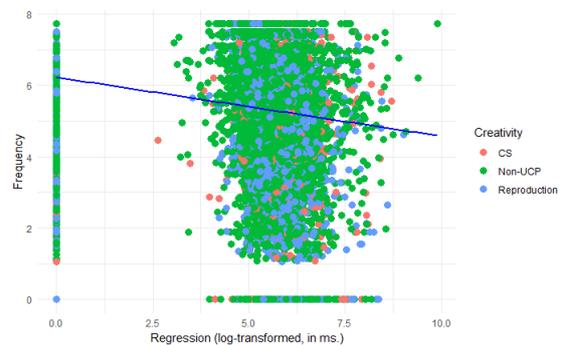}
    \caption{Scatter plot of word frequency and RP (log-transformed). Colour indicates Creativity annotation, showing no clear trend.}
    \label{fig:SP_RP}
\end{figure}

Figure \ref{fig:SP_FPT} shows the scatter plot for word frequency and FPT (log-transformed). We see a similar picture as what we saw in Figure \ref{fig:Frequency_corr}, with no clear trend for Creativity across word frequency. Spearman's correlation shows again a low negative correlation ($\rho$ = 0.23)--even lower than for TFD; this too is highly significant (p < 0.0005). We again see that though there is some relation between FPT and word frequency this is only a low correlation, so there seems to be other factors influencing FPT as well.

Figure \ref{fig:SP_RP} shows the scatter plot for word frequency and RP (log-transformed). We see a similar picture as what we saw in Figure \ref{fig:Frequency_corr}, with no clear trend for Creativity across word frequency. Spearman's correlation shows again a low negative correlation ($\rho$ = 0.29)--similar to TFD; this too is significant (p < 0.0005). We again see that though there is some relation between RP and word frequency this is only a low correlation, so there seems to be other factors influencing RP. 

So, for our three continuous dependent variables (TFD, FPT, and RP) we see a low negative correlation with word frequency. This means that there is a link between word frequency and cognitive load measured, but as this is low, there seems to be other factors influencing the cognitive load exerted. 

\section{Appendix: Analysis of the retrospective think-aloud interviews (RTA)}
\label{sec:appendix-RTA}
This appendix contains the more detailed analysis of the retrospective think aloud interviews. Due to constraints of space, only a quick overview of the results per modality are discussed in the main body of the article. Here we present a more detailed analysis of the themes. The interviews were all coded by one of the researchers after which emerging themes were observed and nodes merged across consecutive coding cycles. In the end, five main themes emerged from the analysis.
\begin{enumerate}
\item Confusion came from the narrative in HT, but from language use in MT 
\item Engaging with and relating to narrative elements occurred in HT, ST \& PE 
\item HT participants felt immersed in the story, the narrative, and the style 
\item MT participants had difficulty understanding the text due to nonsensical words phrasing
\item PE participants were engaged in the narrative, but struggled with the style and characters at times 
\end{enumerate}

\vspace{3mm}

\textbf{1. Confusion came from the narrative in HT, but from language use in MT}

One of the very noticeable things is that across modalities all participants mentioned feeling confused multiple times throughout the narrative: feelings of confusion were mentioned 30 times in HT, 37 times in MT, 28 times in PE, and 23 times in ST. As expected, feelings of confusion were mentioned more in MT, but HT and PE follow closely behind. However, when looking into the reasons participants mention feeling confused, a much clearer image arises: confusion in HT refers mostly to narrative events such as the setting at the beginning and the plot twist towards the end; for MT, however, participants mentioned feeling confused largely related these feelings to words and phrases that were translated incorrectly, difficult to understand or otherwise incompatible with the context. This even made the participants laugh throughout the interview because the words “were just so weird” (P02\_MT). Some clear examples in MT mentioned by both participants were \textit{stripteasenummer} ("striptease number") for “stripling”, \textit{tripjes} (potentially diminutively morphological interpretation of the English original but also meaning "small / short trips") for “triplets”, and \textit{mooie jus} a literal translation of “good gravy”. Other issues in MT included not understanding whole sentences, descriptions, or settings, due to issues with the syntax or simply too many errors. P07\_MT also did not understand what was happening with the painting that one the main characters was painting throughout the story. This confusion was caused in part by the translation of “image” as \textit{foto} ("photo") on multiple occasions while the “image” was in fact referring to a painting. P02\_MT mentioned multiple times that she got confused about the syntax, such as in the sentence \textit{Zie je hier een lijk zonder gezicht waar je me graag met je hoofd op zou willen steken?} (literally "Do you see here a (dead) body without face where you would like to put me with your head on?").

This is in clear contrast with the confusion the HT participants felt. P03\_HT mentioned feeling confused about the title "2BR02B" at first, which is indeed only explained later on in the story as a phone number people can call. She also felt confused about the setting and world building, but this was due to the narrative structure of the story rather than not understanding the words and phrases used--she also got into the story really quickly after reading: on the fourth page, she mentioned "Okay, I see where this is going now" in relation to her previous confusion about the setting of the story. P06\_HT mentioned being a little confused about all the nicknames of the extermination service at first, but feeling engaged in the oppressive atmosphere of the story when discovering they were all “happy-sounding nicknames for suicide machines”. Both also mentioned being confused about the word \textit{steenvruchtje} ("little stone fruit") for “drupelet”, which occurred in a metaphor comparing the overcrowded world to a stone fruit-—although it is highly imaginable that people would have been confused about the original ‘drupelet’ as well, given that it occurs fewer than 0.01 times per million words in modern English according to the OED (“drupelet”, n.1). This seems to be the case indeed as the ST participants also mentioned feeling confused about the "drupelet" the metaphor surrounding this: "I had to think about this, it takes a while to get it, it's because this [the drupelet, \textit{red.}] got me a little confused and then, I mean what I know what he means" (P04\_ST). For HT and ST then, confusion came mostly from narrative choices or lexical choices that were similar to the original (such as the long enumeration of nicknames, which confused both P04\_ST and P03\_HT), while for MT confusion was caused by errors and translation issues.

\vspace{3mm}

\textbf{2. Engaging with and relating to narrative elements occurred in HT, ST \& PE}

HT, ST and PE participants mentioned feeling engaged in the narrative, including the events of the story and the moral issues at play, relating it to their own lives often. For these modalities, participants mentioned that the story was interesting, with a well set up moral dilemma, making them empathise with Edward’s choices. P06\_HT specifically imagined how she herself would react to living in a world like that. In general, both HT participants mentioned they felt very immersed in the story, liking the characters (the second participant really loved the painter, saying he was going to be one of her new favourite characters) and describing how each character had a very distinctive style and feel to her. P08\_ST mentioned he would like to read more stories by this author and that he felt very engaged in the narrative and the style: "I was also very curious to see what would happen next" and that he felt bad for the main characters too. P04\_ST also said that the text was "nice to read". P09\_PE mentioned really liking some of the “strong imagery” created in the story, such as the image of Leora, the word \textit{Kattenbak} ("Catbox”) for the suicide chambers and the image of \textit{dompelen} ("dunking") people as a kind of baptism—even though he also felt that the word itself was not used completely correct: "The painter started talking about like "baptising", \textit{dompelen} ("baptising"), and I was thinking that it was a very euphemistic description. I don't know, I thought it was interesting (...) but I also felt like it came out of nowhere and that it didn't really fit, at least not in the way it intended to".  P04\_ST also mentioned multiple times that she thought descriptions were chosen well and felt fitting in the story and the setting. The participants also mentioned how some of the elements of the story made them think of their own lives and experiences. P05\_PE, for instance, mentioned how the description of the colour purple as "the color of grapes on Judgment Day" made her think of the art in the Galleria Borghese and the description of the character of 
Leora of her own mother. P09\_PE mentioned relating the story to the Second World War and trying to recognise the song and creating a little melody to go with it. HT participants also related the story to their situation, with P06\_HT relating the society in the story and specifically the description of the world as it was in the narrative past to the current Dutch society; she also liked the reference to Zeus in the text as a Classics’ enthusiast. The other (P03\_HT) started talking about a pin she herself had bought for a friend of hers which resembled Leora’s pin, and towards the end, how making the appointment for the suicide chambers resembled making an appointment at the dentist. This was not the case for MT. P02\_MT did not relate the text to her own life, only relating some of the in their eyes more surprising errors to text-external things: she linked the \textit{grove vrouw} ("coarse woman") to \textit{grove mosterd} ("coarse-grained mustard") and \textit{ontlasten} (litt. "relieve") to peeing rather than "disposing of someone" as in the original. P07\_MT did not relate any part of the story to her own life at all, only mentioning how weird things sounded or how it should have been in Dutch to reconstruct the story (e.g. "'dompelen mensen onder' ("immerse people") I didn't completely understand but it's probably about people who are dying" or "'oude eend' ("old duck") was also funny, like okay, 'old man' I'd say or 'oude lul' (litt. "old dick") or something"). Research has shown that readers who relate parts of a story to their own experiences and own frames of understanding and seeing the world feel more engaged and like a story better (Kuiken et al., 2004). 
\vspace{3mm}

\textbf{3. HT participants felt immersed in the story, the narrative, and the style }

Throughout the interviews, the HT participants made it clear that they liked the story in many of its facets and felt immersed in both the narrative and the style. P06\_HT kept commenting about how immersed she felt in the story, how much she liked the character of the painter, how much she empathised with Edward and how well-put the moral dilemma was. She specifically mentioned enjoying the dialogue and the “ironic and witty” banter back and forth between the painter and the nurse, describing the dialogue with Dr. Hitz from both perspectives in the story, seeing Dr. Hitz’s appreciation of the system but also the “painful” decision for Edward. She also mentioned multiple comical instances in the story, such as Leora’s moustache and the way she and the painter squabbled about which figure fit her best. She also described how she would feel and act from the different characters’ viewpoint, clearly immersing herself and placing herself in the story. P03\_HT also engaged emotionally with the characters, describing Dr. Hitz “as being just so annoying, [which] works so well for the story”. She described the story as “engrossing” and mentioned how the story kept a good balance between explaining and showing the world-building. She also pointed out the wordplay in \textit{Duncan} and \textit{dunken} ("dunking"), which she felt was not only very good and expressive, but also a good find on the translator’s part. Both mentioned how the story made them think about the world, the story world, what they would do themselves in such a situation and whether the story world is a better world than our current world. Still there was also some confusion in the HT version, but this tended to be related to the narrative and fit general reading experiences (especially for short stories), such as confusion about world-building at the beginning and surprise at plot twists. The HT participants engaged deeply with the story, feeling immersed in the narrative and the characters, appreciating the style and the way the moral dilemma made them reconsider some of the values and situations in the world.
\vspace{3mm}

\textbf{4. MT participants had difficulty understanding the text due to nonsensical words phrasing}

Translation errors in MT led to nonsensical phrasings, which caused the participants to struggle understanding the narrative and its events. Participants mentioned that they were not sure what was happening at multiple times during the RTAs ("[I] just didn't really see what was happening here" (P02\_MT) \& "I couldn't follow what it said" (P07\_MT)). This made it difficult for them to retain and envision the story in their minds, including which character was which, what their role and potential development was in the story and what had happened so far in the narrative: "I couldn't really connect with the characters here, she seemed very, yeah, I don't know, problematic? But yeah, it's also just like the text, you know the words and stuff, feel like there's a barrier there or something… also because I just don't really know what's actually there or what's weird or something" (P02\_MT). Participants also mentioned struggling with retaining the developments of the plot in mind as  they were continuously trying to reconstruct the ‘correct version’ of the text and events in their mind while reading. P07\_MT mentioned “it was funny; you know what they are trying to say, but it does not work like that, and it is definitely not correct.” P02\_MT also described trying to “reconstruct” the correct version of the text in her mind, but she said that this made her feel very detached from the story and caused her to, at times, read the text cursory rather than in-depth because “[she] had no idea what was going on anyway”. Rather than trying to reconstruct the narrative, she also mentioned giving up at times:  "I also think here is kind of where I also started giving up? Or like, not necessarily actually giving up but more like, accepting that I wouldn't really get the thing". When discussing the metaphorical image of the drupelet, she also mentioned not even bothering to recreate the image in her mind, because she did not believe she would understand the metaphor anyway.

However, this does not mean participants hated the story. Both mentioned liking certain parts of the narrative. One of the participants felt the ending was very fitting for the story and mentioned liking the moral dilemma, describing the story as a “gripping sci-fi story” (P07\_MT). Both also mentioned that they believed they would like the story a lot more in English ("[I think] I'd prefer to English original" (P02\_MT). Still, it is clear that on the word and stylistic level, MT was strongly inadequate, obfuscating understanding of the text and even for those sections where the meaning could be reconstructed making readers feel detached and disengaged from the different story elements and the plot as a whole.
\vspace{3mm}

\textbf{5. PE participants were engaged in the narrative, but struggled with the style and characters at times}

PE participants liked the story overall, thought it set-up the moral dilemma really well, and enjoyed themselves while reading the story. Both participants related the situation and parts of the setting in the story to their own lives and experiences, and one of the participants specifically mentioned the “strong imagery” (P09\_PE) in the story throughout. When PE participants expressed their confusion, this tended to be related to the narrative elements in a similar way as the HT participants’ confusion, rather than any confusion caused by nonsensical phrasing or other (blatant) translation errors as happened for MT participants. At the same time, however, PE participants did not like the style: P09\_PE, who was a little milder than the other, said that the style “did not struck [him] specifically”, but liked it well enough, although he also mentioned that “some sentences seemed off, not specifically clear why but the phrasing seems off”. P05\_PE was more forceful and negative about the style, saying that “[she] realised it had to be a translation, because no Dutch person would have written this like this”. However, both participants found it difficult to exactly pinpoint instances in which they disliked the writing. This could be caused by the fact that there were almost no glaring errors in the text (which MT did have), but rather just a general feeling of the text not adhering to normal Dutch writing styles. The participants did mention some adjectival use that felt strange and some of the words that seemed to be out of context. The instance of \textit{broeder} ("brother") for the nurse is discussed above, in which the chosen translation is not so much incorrect, but rather uncommon and more commonly used in other contexts (monks or in the rap and street scene colloquially). It is possible that there were more of such instances, which were less conspicuous but influenced the reading experience. 

One of the other surprising things that happened with both PE participants is that they confused multiple characters. P09\_PE confused Edward (the father) with the painter, while P05\_PE confused Edward with Dr. Hitz. It is true that the story does not have a clear main character, with all three playing an important role in different parts of the story, but it is noticeable that both participants had issues with keeping the characters apart. This was also not just a brief confusion of characters, but both participants only realised their error during the questionnaire when the multiple-choice options included all characters. It is a little unclear what caused this confusion. Both participants mentioned that characters’ motivations were not always clear, although P09\_PE said he liked the characterisation overall. P05\_PE was more critical about the characterisation, feeling that it was done “rather poorly”, with characters’ emotions shifting immensely without any explanation or emotions she could not place in general, also mentioning that she “couldn’t really connect to the characters”. P09\_PE did comment that the adjectival use was weird throughout the story, especially pointing to the adjectives that were used to describe characters, such as \textit{een grimmige oude man} ("a grim old man") \textit{een grove (…) vrouw} ("a coarse woman"),  and P05\_PE also mentioned feeling confused about the description of the hospital brother as \textit{broeder}, which in Dutch is acceptable but not very commonly used. Both also mentioned that they felt shifts in the story were very sudden and that the different sections were not well connected.

Lastly, PE participants seemed more confused (and for a longer period of time) than HT participants: like the HT participants, both PE participants mentioned feeling confused about the image of the drupelet; however, PE participants seemed to understand the imagery only later during the RTAs, while HT participants said they understood it almost directly when reading the text for the first time. P05\_PE also mentioned not fully understanding the ending: “Everyone dies and then you have the painter, and he continues to paint or something?”; although this also relates to the narrative level as the confusion in HT did, it seemed that in PE these confusing elements were not always solved (as they were in HT), which left PE readers with a lower appreciation (as shown in the RTAs) and potentially comprehension (as shown in the questionnaire) than HT readers had for these narrative elements. 

Interestingly, it were also the PE participants who had the lowest score for comprehension in the questionnaire (a mean score of 6.5, see Table \ref{table:questionnaire})--this relative low comprehension could be linked to the confusion the participants mentioned in the RTAs. A potential cause for this confusion and lower comprehension in PE are the lack of connections and particles in the PE version. PE participants mentioned that they felt shifts in the story were very sudden and that the different sections were not well connected: "it [the narrative, \textit{red.}] seemed to jump around in like the setting and characters and like the shifts from one thing to the next were a bit inconclusive, or random". These sudden jumps and shifts could be caused by the lack of connectors and particles in PE, which are typical of Dutch language and studies have shown PE struggles with these at times (\citealp{Kroon2023}; \citealp{Lefer2021}). This could cause the lack of cohesion felt by the PE participants which in turn could potentially explain the lower comprehension of PE participants throughout. However, it could also be that the specific PE participants just struggled more with the text or that for these two participants the experimental conditions (such as reading from a computer while resting their head in a headset) had more impact on their general reading experience. 
\end{document}